\documentclass{article} % For LaTeX2e
\usepackage{iclr2027_conference,times}

\usepackage{amsmath,amsfonts,bm}

\def\eqref#1{equation~\ref{#1}}
\def\1{\bm{1}}

\DeclareMathAlphabet{\mathsfit}{\encodingdefault}{\sfdefault}{m}{sl}
\SetMathAlphabet{\mathsfit}{bold}{\encodingdefault}{\sfdefault}{bx}{n}

\newcommand{\R}{\mathbb{R}}

\usepackage{hyperref}
\usepackage{graphicx}
\usepackage{url}
\usepackage{amsfonts, amsmath}       % blackboard math symbols
\usepackage{amsthm}
\usepackage{enumitem}

\usepackage{algorithm}
\usepackage{algpseudocode}
\usepackage{booktabs}

\theoremstyle{plain} 

\newtheorem{definition}{Definition}
\newtheorem{hypothesis}{Hypothesis}

\title{Converge to Surprise: Evolutionary Self-supervised Image Clustering}

\author{Canlin Zhang\thanks{This work is supported by Sorensom Communications AI-Lab. Code available at \url{https://github.com/canlinzhang/converge-to-surprise}.}\\
Independent Researcher\\
{\tt\small canlingrad@gmail.com}
\and
Xiuwen Liu\\
Department of Computer Science\\
Florida State University\\
{\tt\small xliu@fsu.edu}
}

\preprint % Uncomment for a pre-print version
\begin{document}

\maketitle

\begin{abstract}
A variety of self-supervised image clustering approaches are invented in the past years. However, all dominant approaches are \emph{exploitative}: The direction of parameter updates is determined by known states (observed input samples and existing parameters). We propose an \emph{explorative} self-supervised learning framework that steps out of this zone. We define a \emph{surprise score} that measures how unlikely the model's output representation is, assuming that all pixels are i.i.d.\ random noise. Maximizing the surprise score forces the deep learning model to reject the random noise null hypothesis, or equivalently, to discover \emph{non-randomness} from data. Also, we propose a fundamental assumption: a surprise score cannot, in general, be fully optimized by exploitative optimization approaches. Thus, we propose the \emph{converge-to-surprise} scheme to optimize a model: an evolution-strategy (ES) outer loop, which maximizes the surprise score using the mutation-selection mechanism, paired with a periodic gradient-descent inner loop, which uses the surprising clusters already discovered by ES as surrogate targets. On simple image benchmarks, our framework trained from scratch achieves new state-of-the-art results in non-parametric self-supervised image clustering --- the strictest deep-clustering setting, where the number of classes is unknown during training.
\end{abstract}

\section{Introduction}
\label{sec:intro}
%In recent years, generative AI, especially large language models (LLMs) \citep{brown2020language, achiam2023gpt}, has achieved striking success across a wide range of tasks. The core principle to train an LLM is amazingly simple: Predicting the next token \citep{AttentionIsAll, radford2018improving}, or predicting masked tokens \citep{Devlin_BERT} in the context. However, when it comes to raw vision (pure unlabeled images and videos), there is no token or \emph{symbolic} representation available any more. Hence, self-supervised learning from raw vision data has not kept pace with the success in LLMs.

Self-supervised image clustering aims at grouping unlabeled images into distinct, semantically meaningful categories without human intervention \citep{chen2020simple, he2022masked}. This is achieved by unifying self-supervised representation learning and unsupervised clustering into a cohesive pipeline \citep{caron2018deep, caron2021emerging}.

A long line of work has approached this problem with different efforts. Patch-arrangement methods predict structural corruptions: a jigsaw permutation \citep{noroozi2016unsupervised}, an applied rotation \citep{gidaris2018unsupervised}, or the relative position of two crops \citep{doersch2015unsupervised}. Contrastive learning pulls together two augmented views of one image and pushes apart views of different images \citep{chen2020simple, he2020momentum}. Masked autoencoders reconstruct randomly masked pixels \citep{he2022masked}. Deep clustering alternates between training on pseudo-labels and re-clustering the output features \citep{caron2018deep, caron2020unsupervised}. Self-distillation trains a student to match a teacher whose weights are an exponential moving average of the student's \citep{grill2020bootstrap, caron2021emerging}. Latent-space prediction, the joint-embedding-predictive-architecture (JEPA) family, predicts the embedding of one part of an image from that of another \citep{assran2023self}. Information-theoretic methods optimize differentiable surrogates for mutual information between views, or for de-correlation across feature dimensions \citep{zbontar2021barlow, bardes2021vicreg, hjelm2018learning, ji2019invariant}.

In all these approaches, impacts made by random perturbations (such as data augmentation \citep{perez2017effectiveness} and dropout \citep{srivastava2014dropout}) are only up to an incidental level: There is no mutation-selection mechanism \citep{bell2008selection} that compares the model's performances under different perturbations, and then updates the parameters based on the results. The loss, or in general the direction of parameter updates, is determined by the observed samples in the current step and the model parameters in the current stage. That is, all these approaches are \emph{exploitative} \citep{hill1993exploitation}.

%In all these approaches, the loss, or in general the direction of parameter updates, is determined by the observed samples in the current step as well as the model parameter in the current stage. Impacts made by random perturbations (such as data augmentations \citep{perez2017effectiveness} and dropout \citep{srivastava2014dropout}) are only up to incidental level: There is no mutation-selection mechanism that compares the model's performances under different perturbations and then update the parameters based on the comparison. That is, all these approaches are \emph{exploitative} \citep{hill1993exploitation}.

%given the observed samples in the current step, the training targets (structural corruptions, masked regions, contrastive signals, clustering results, outputs of an EMA-updated teacher, etc) are obtained using the model's current parameters. Furthermore, the current parameters of the model are updated based on the samples and parameters in previous steps. Thus, after initialization, the deep learning model is shaped and determined by the data it encounters, which is a \emph{exploitative} process \citep{hill1993exploitation}.

In contrast, this paper introduces a self-supervised learning (SSL) framework that applies a mutation-selection mechanism to parameter updates, leading to an \emph{explorative} \citep{he2004exploration} image clustering approach. Initially, we ask one question: \emph{what is not random in an image?} That is, in the absence of prior knowledge, the Principle of Maximum Entropy \citep{max_entropy} dictates the most conservative assumption on the image distribution — pixels are i.i.d.\ random noise. We take this assumption as our null hypothesis $\mathcal{H}_0$, and define a \emph{surprise score} that measures how unlikely the model's output representation should occur under $\mathcal{H}_0$. The higher the surprise score, the less plausible $\mathcal{H}_0$ becomes. Thus, maximizing the surprise score forces the model to reject the null hypothesis $\mathcal{H}_0$. Equivalently, this makes the model extract meaningful information, or \emph{non-randomness}, from images. We show that maximizing the surprise score naturally prevents representation collapse. Finally, the obtained representation is used for image clustering.

Then, by introducing formal definitions for exploitative and explorative SSL approaches, we propose a fundamental hypothesis of this paper: \emph{a surprise score cannot, in general, be fully optimized by exploitative optimization approaches}. Although proving this hypothesis is beyond our scope, detailed analysis is provided.

Accordingly, we propose the \emph{converge-to-surprise} scheme: An evolution-strategy (ES) \citep{salimans2017evolution} outer loop, which maximizes the surprise score by mutation-selection mechanism, is paired with a periodic gradient-descent inner loop, which uses the surprising clusters already discovered by ES as surrogate targets. Experiments on simple image datasets (MNIST \citep{MNIST_paper}, FashionMNIST \citep{xiao2017fashion} and USPS \citep{hull2002database}) show that our scheme achieves new state-of-the-art results in non-parametric self-supervised image clustering — the strictest deep-clustering task, in which case the number of ground-truth classes is unknown to the model during training.

Here are our contributions:
\begin{enumerate}[leftmargin=*]
\item We define a surprise score that measures how unlikely the output of a deep learning model should occur, under the random noise null hypothesis. Maximizing the surprise score makes the model discover non-randomness from data, which naturally prevents representation collapse. Then, we formally define whether an SSL approach is exploitative or explorative. We show that all dominant SSL approaches for image clustering are exploitative. Finally, we propose a fundamental hypothesis: a surprise score cannot, in general, be fully optimized by exploitative approaches.
\item We propose converge-to-surprise, an explorative optimization scheme that combines evolution strategy with gradient-descent training to maximize the surprise score.
\item On simple image benchmarks, models trained from scratch using our scheme achieve new state-of-the-art results in non-parametric self-supervised image clustering tasks --- the strictest image clustering scenario. To the edge of our knowledge, our model is the first explorative SSL approach being trained from scratch for self-supervised image clustering. 
\end{enumerate}
The remainder of the paper is organized as follows: Section~\ref{sec:related-work} surveys related work in self-supervised image clustering. Section~\ref{sec:main-theory} describes the converge-to-surprise framework with all details. Section~\ref{experiments} reports experimental results, and Section~\ref{sec:conclusion} concludes.

\section{Related Work}
\label{sec:related-work}

We organize self-supervised image clustering methods by what the network outputs, and how that output becomes a cluster assignment at test time.

\textbf{Embedding-based clustering}: Most self-supervised image clustering methods produce a dense feature embedding in $\mathbb{R}^d$, which is further clustered or linearly probed at test time. These span contrastive methods (SimCLR \citep{chen2020simple}, MoCo \citep{he2020momentum}), self-distillation (BYOL \citep{grill2020bootstrap}, DINO \citep{caron2021emerging}), latent-space prediction (I-JEPA \citep{assran2023self}), masked autoencoders \citep{he2022masked}, information-theoretic objectives (Barlow Twins \citep{zbontar2021barlow}, VICReg \citep{bardes2021vicreg}, Deep InfoMax \citep{hjelm2018learning}), and pseudo-label clustering (DeepCluster \citep{caron2018deep}, SwAV \citep{caron2020unsupervised}). They differ in the training signal but agree on the output: a continuous embedding, not a hard cluster assignment.

These approaches first apply a deep encoder that maps images to a continuous embedding space. Then, a shallow decoder (usually a multi-layer perceptron \citep{popescu2009multilayer}) maps the embedding to the output representation. In test time, classification is made based on nearest-neighbor evaluation \citep{mucherino2009k} or via fine-tuned linear projections using the embeddings \citep{caron2018deep}. This is the most dominant yet mildest setting: The network is not required to produce hard representations of the image, only an embedding that captures semantic meaning.

\textbf{Parametric hard clustering}: A second family outputs a cluster index directly, mapping each input to one of $K$ clusters with $K$ equals to the number of true classes. At evaluation, clusters are matched to class labels by the Kuhn-Munkres (Hungarian) algorithm \citep{kuhn1955hungarian}. DEC \citep{xie2016unsupervised} sharpens a soft assignment over $K$ centroids, and DAC \citep{chang2017deep} recasts clustering as pairwise same/different classification. SCAN \citep{van2020scan} decouples feature learning from cluster assignment by classifying images based on semantic nearest neighbors. CLUDI \citep{uziel2025clustering} distills stochastic diffusion-based cluster assignments from a teacher into a student, on top of a pretrained DINO model \citep{caron2021emerging}. Although being a parametric approach, IIC \citep{ji2019invariant} is the closest one to our framework, which maximizes softmax mutual information \citep{kraskov2004estimating} between the cluster predictions of two augmented views. 

\textbf{Non-parametric hard clustering}: The third family outputs a cluster index without knowing the number of true classes in advance. This is the strictest image clustering regime, and the setting for which we report results. Few enough approaches can be found in this family. DeepDPM \citep{ronen2022deepdpm} changes $K$ during training by split/merge operations inspired by Dirichlet-Process Mixture (DPM) models \citep{neal2000markov}. UNSEEN \citep{leiber2024dying} applies existing deep-clustering backbones (DCN \citep{yang2017towards}, DEC \citep{xie2016unsupervised}, DKM \citep{fard2020deep}), then starts from an upper bound $K_{\max}$ and lets unused clusters atrophy. %The deep DPM model of \citep{li2022deep} combines a flow-based network with Gibbs sampling over an infinite-component DPM prior. 
Classical algorithms such as moVB \citep{hughes2013memoized} and DBSCAN \citep{ester1996density} operate on learned features. All these models serve as baselines in this paper.

\section{Main Theory}
\label{sec:main-theory}

In this section, we first introduce our complementary masking strategy, based on which the surprise score is calculated. Then, we formally define exploitative and explorative optimization approaches. Finally, we combine evolution strategy with gradient-descent to produce our explorative approach.

\subsection{Complementary masking strategy}
\label{sec:masking}
Suppose we have a distribution $\mathbf{P}$ producing images of shape $(H, W, C)$. That is, $\mathbf{X} \sim \mathbf{P}$ with $\mathbf{X} \in \mathbb{R}^{H, W, C}$. In a self-supervised learning scenario, we have no prior knowledge about $\mathbf{P}$. According to the Principle of Maximum Entropy \citep{max_entropy}, we have to assume that each pixel $(h,w)$ in $\mathbf{X}$ is independent of every other. In other words, we have to assume that $\mathbf{P}$ is the maximum-entropy distribution over $\mathbb{R}^{H, W, C}$, producing totally random noise. This is our null hypothesis $\mathcal{H}_0$.

%If $\mathcal{H}_0$ is correct, we have no reason to further implement a deep learning model on samples of $\mathbf{P}$. From this point of view, the purpose of self-supervised learning is to optimize a deep learning model so that its output representation rejects the random noise null hypothesis, and discovers \emph{non-randomness} from data.

%There are many methods to evaluate whether a model's output rejects the null hypothesis $\mathcal{H}_0$. Here, we introduce one method that we think is straightforward:

\begin{figure}[h]
    \centering
    \includegraphics[width=0.45\linewidth]{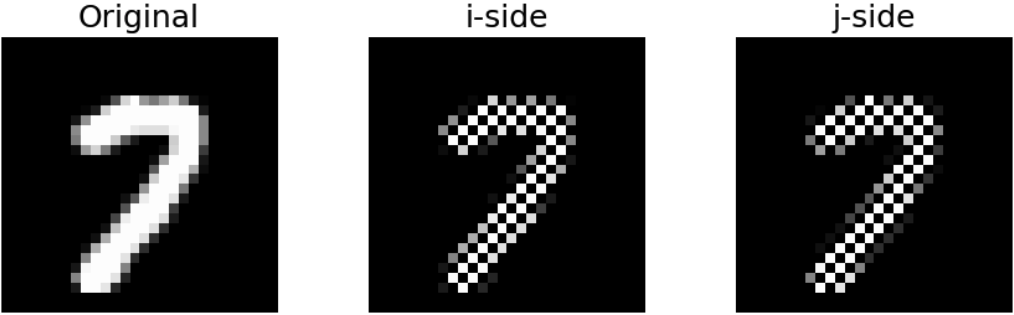}
    \caption{Chessboard masking: $i$-side can only view pixels in white chessboard positions; $j$-side can only view pixels in black chessboard positions. The two views share no pixel in common.}
    \label{fig:masking}
\end{figure}

Then, given a sampled image $\mathbf{X} \sim \mathbf{P}$, we partition its $H \times W$ pixel grid into two disjoint subsets using a \emph{chessboard pattern}. Let
\begin{equation}
    \mathcal{I} = \{(h, w) : (h + w) \bmod 2 = 0\}, \ \ \ \ \mathcal{J} = \{(h, w) : (h + w) \bmod 2 = 1\}
\end{equation}
denote the white and black chessboard positions, respectively. Let $\mathbf{M}^{(\mathcal{I})}, \mathbf{M}^{(\mathcal{J})} \in \{0, 1\}^{H, W}$ be the corresponding binary masks, defined by $\mathbf{M}^{(\mathcal{I})}_{h, w} = \mathbf{1}\!\left[(h + w) \bmod 2 = 0\right]$ and $\mathbf{M}^{(\mathcal{J})} = \mathbf{1} - \mathbf{M}^{(\mathcal{I})}$, where $\mathbf{1}[\cdot]$ is the indicator function \citep{mclean1991indicator}. Applying these masks pixel-wise to $\mathbf{X}$, we obtain two complementary masked views of the same image:
\begin{equation}
    \mathbf{X}^{(i)} = \mathbf{X} \odot \mathbf{M}^{(\mathcal{I})}, \quad \mathbf{X}^{(j)} = \mathbf{X} \odot \mathbf{M}^{(\mathcal{J})},
\end{equation}
where $\odot$ denotes element-wise multiplication. The $i$-side view $\mathbf{X}^{(i)}$ retains only the pixels at white chessboard positions and zeros out the rest, while the $j$-side view $\mathbf{X}^{(j)}$ retains only those at black positions. Every pixel of $\mathbf{X}$ appears in exactly one of the two views, and the two views share no pixel in common. Figure~\ref{fig:masking} shows our masking strategy on an MNIST image \citep{MNIST_paper}.

As mentioned, the null hypothesis $\mathcal{H}_0$ assumes pixels of $\mathbf{X}$ to be independent of one another. Since $\mathbf{X}^{(i)}$ and $\mathbf{X}^{(j)}$ share no common pixel, they are themselves independent tensors and therefore share no mutual information \citep{mutual_information}. In addition, applying data augmentation \citep{perez2017effectiveness} independently on $\mathbf{X}^{(i)}$ and $\mathbf{X}^{(j)}$ will not break this zero-mutual-information statement. We use $\tilde{\mathbf{X}}^{(i)}$ and $\tilde{\mathbf{X}}^{(j)}$ to denote the augmented views. Therefore, we have $ \mathcal{H}_0 \Longrightarrow I\!\left(\tilde{\mathbf{X}}^{(i)} \,;\, \tilde{\mathbf{X}}^{(j)}\right) = 0$, where $I(\cdot \,;\, \cdot)$ denotes the mutual-information functional. 

If $I\!\left(\tilde{\mathbf{X}}^{(i)} \,;\, \tilde{\mathbf{X}}^{(j)}\right) = 0$ is rejected, then $\mathcal{H}_0$ cannot be true. This will be our approach to reject $\mathcal{H}_0$.

\subsection{Cluster co-occurrence as a surprise score}
\label{sec:surprise}

Again, given an image $\mathbf{X} \sim \mathbf{P}$, we obtain the two complementary masked and independently augmented views $\tilde{\mathbf{X}}^{(i)}$ and $\tilde{\mathbf{X}}^{(j)}$. Then, a deep learning model $f_\theta : \mathbb{R}^{H \times W \times C} \to \mathbb{R}^{K}$ assigns each view to one of $K$ candidate clusters, according to the argmax dimension in its output logits:
\begin{equation}
    \hat y^{(i)} = \arg\max_{k}\bigl[f_\theta(\tilde{\mathbf{X}}^{(i)})\bigr]_k,
    \quad
    \hat y^{(j)} = \arg\max_{k}\bigl[f_\theta(\tilde{\mathbf{X}}^{(j)})\bigr]_k.
    \label{eq:argmax_logit}
\end{equation}
Both $\hat y^{(i)}$ and $\hat y^{(j)}$ take values in $\{0, 1, \ldots, K-1\}$. Because each $\hat y^{(\cdot)}$ is a deterministic function of the corresponding view, the data-processing inequality \citep{beaudry2011intuitive} gives
$I(\hat y^{(i)};\,\hat y^{(j)})\le I\!\left(\tilde{\mathbf{X}}^{(i)} \,;\, \tilde{\mathbf{X}}^{(j)}\right) = 0$
under $\mathcal{H}_0$. That says, the cluster pseudo labels from $i$-side and $j$-side are statistically independent.

Then, given $N$ images $\{\mathbf{X}_1, \dots, \mathbf{X}_N\}$ sampled i.i.d.\ from $\mathbf{P}$, we obtain their complementary masked and independently augmented views $\{\tilde{\mathbf{X}}_1^{(i)}, \dots, \tilde{\mathbf{X}}_N^{(i)}\}$ and $\{\tilde{\mathbf{X}}_1^{(j)}, \dots, \tilde{\mathbf{X}}_N^{(j)}\}$, respectively. We implement our model $f_\theta$ on each view and obtain the predicted cluster, leading to two integer sequences of length $N$. That is, $    \mathrm{seq}^{(i)} = \bigl(\hat y_1^{(i)}, \dots, \hat y_N^{(i)}\bigr)$ and $ \mathrm{seq}^{(j)} = \bigl(\hat y_1^{(j)}, \dots, \hat y_N^{(j)}\bigr)$.

In fact, we implicitly assume that the distributions of nearby pixels in an original image $\mathbf{X}$ are almost identical, although pixels are assumed to be i.i.d.\ under $\mathcal{H}_0$. Hence, after chessboard masking, $\mathbf{X}^{(i)}$ and $\mathbf{X}^{(j)}$ will have identical distribution. After independent augmentation with the same hyper-parameters, $\tilde{\mathbf{X}}^{(i)}$ and $\tilde{\mathbf{X}}^{(j)}$ will also have identical distribution. Thus, although statistically independent, $\mathrm{seq}^{(i)}$ and $\mathrm{seq}^{(j)}$ will have identical distribution as well under $\mathcal{H}_0$.

For each cluster $k \in \{0, 1, \dots, K-1\}$, let $n_k^{(i)}$ be the number of times $k$ appears in $\mathrm{seq}^{(i)}$, and define $n_k^{(j)}$ analogously. Then, the empirical marginal probability of clustering a view (either $i$-side or $j$-side) to cluster $k$ is $p_k \;=\; (n_k^{(i)} + n_k^{(j)})\ /\ (2N)$.

Again, since $\mathrm{seq}^{(i)}$ and $\mathrm{seq}^{(j)}$ are independent under $\mathcal{H}_0$, the null probability that both views of one image are predicted to the same cluster $k$ (i.e. $\hat y_n^{(i)} = \hat y_n^{(j)} = k$ at any index $n$) is $q_k \;=\; p_k \cdot p_k \;=\; p_k^{\,2}$. Hence, under $\mathcal{H}_0$, the expected number of view matching at cluster $k$ over the whole batch is $N\cdot q_k$. 

On the other hand, we denote the observed number of view matching at cluster $k$ as $t_k$, which is the actual count of indices $n \in \{1, \dots, N\}$ for which $\hat y_n^{(i)} = \hat y_n^{(j)} = k$. Accordingly, the observed view-matching probability is $\hat q_k = t_k / N$.

Considering cluster $k$ in isolation, $t_k$ is then the number of successes in $N$ independent Bernoulli trials, each with success probability $q_k$ \citep{binomial_distribution}. That is, $t_k \sim \mathrm{Binomial}(N,\, q_k)$ under $\mathcal{H}_0$.

The $K$ output dimensions are of course coupled: each view is predicted to exactly one cluster. So, the counts $t_1, \dots, t_K$ are correlated. But this coupling has only a very weak effect on the distribution of any individual $t_k$. Treating each $t_k$ on its own as binomial is the form we will use throughout. 

Then, we apply the binomial distribution formula \citep{feller1991introduction}:
\begin{equation}\label{exact_prob_definition}
    \mathbb{P}_{\mathcal{H}_0}\!\left[t_k \text{ or more matches at } k\right]
    = \sum_{s=t_k}^{N} \binom{N}{s}\, q_k^{\,s}\,(1 - q_k)^{N - s}.
\end{equation}
The sum on the right is exact but awkward to compute. A standard Chernoff upper bound on the binomial upper tail \citep{cover1999elements} turns it into the compact inequality
\begin{equation}\label{prob_surprise}
    \mathbb{P}_{\mathcal{H}_0}\!\left[t_k \text{ or more matches at } k\right]
    \le \exp\bigl(-N\!\cdot\! D(\hat q_k \,\Vert\, q_k)\bigr),
\end{equation}
where $D(\hat q_k \,\Vert\, q_k)$ is the binary Kullback--Leibler divergence \citep{shlens2014notes_KL_divergence_2} between the observed view-matching probability $\hat q_k = t_k / N$ and the null view-matching probability $q_k=p_k^{\,2}$:
\begin{equation}\label{surprise_score_k}
    D\!\left(\hat q_k \,\Vert\, q_k\right)
    \;=\; \hat q_k \log\frac{\hat q_k}{q_k}
    \;+\; (1-\hat q_k)\log\frac{1 - \hat q_k}{1 - q_k}.
\end{equation}
After taking logarithm, formula \ref{prob_surprise} becomes
\begin{equation}\label{log_formula_surprise}
    -\log(\mathbb{P}_{\mathcal{H}_0}\!\left[t_k \text{ or more matches at } k\right]) \;\ge\; N\, D(\hat q_k \,\Vert\, q_k),
\end{equation}
where the left side is the amount of information, or \emph{level of surprise}, obtained by observing the view-matching result at cluster $k$ \citep{shannon1948mathematical_information_theory}. The larger $N\!\cdot\! D(\hat q_k \,\Vert\, q_k)$ is, the more surprising and thus unlikely the result will be under $\mathcal{H}_0$. We call $D(\hat q_k \,\Vert\, q_k)$ the \emph{surprise score at cluster $k$}.

\begin{figure*}[t]
    \centering
    \includegraphics[width=\linewidth]{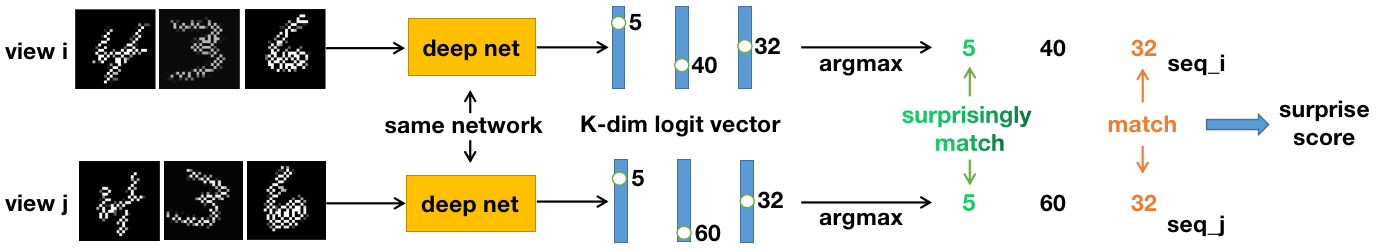}
    \caption{Overview of our pipeline. The deep network is implemented on complementary masked and independently augmented views. Then, argmax is obtained from each output logit vector. Finally, surprise score sums across over-matching clusters.}
    \label{fig:pipeline}
\end{figure*}

Formula \ref{log_formula_surprise} shows the probability of observing $t_k$ or more view matching at cluster $k$ under the null hypothesis $\mathcal{H}_0$. So, we will ignore the surprise score if $\hat q_k < q_k$ for any cluster $k$ (under-matching). Then, summing across all over-matching clusters, we define the final \emph{surprise score}
\begin{equation}
    \mathcal{S}(\theta) \;=\; \sum\nolimits_{k\,:\, \hat q_k > q_k} D\!\left(\hat q_k \,\Vert\, q_k\right).
    \label{eq:surprise}
\end{equation}
$\mathcal{S}(\theta)$ measures in total how strongly the view-clustering results reject $\mathcal{H}_0$, or how surprising the view-matching results look like. To achieve a large $\mathcal{S}(\theta)$, the model needs to discover reliable features shared by both views, which are essentially the meaningful information, or \emph{non-randomness}, in data distribution $\mathbf{P}$. Hence, the optimization goal of our model is to maximize $\mathcal{S}(\theta)$. We call this framework \emph{converge-to-surprise}. Figure \ref{fig:pipeline} briefly illustrates the pipeline of our framework.

If the argmax of all logit vectors collapse to one fixed dimension, then it is easy to see that $\mathcal{S}(\theta)=0$. In contrast, if the pixels are really random noise, we can also have $\mathcal{S}(\theta)\approx 0$ since the two views will be independent. Hence, maximizing $\mathcal{S}(\theta)$ naturally prevents representation collapse \citep{li2022understanding}. Also, the argmax operation makes $\mathcal{S}(\theta)$ irrelative to $K$ in its formula, enabling non-parametric clustering. In Appendix \ref{connect_surprise_score_to_IIC}, we show the connection between our surprise score $\mathcal{S}(\theta)$ and the IIC model training target \citep{ji2019invariant}.

However, optimizing $\mathcal{S}(\theta)$ is not straightforward. We first provide the following definition:

%We admit that the argmax operation makes $\mathcal{S}(\theta)$ non-differentiable to $\theta$. This differs from the softmax mutual information maximization in IIC \citep{ji2019invariant}, which is differentiable end-to-end, yet \emph{parametric} (requiring $K=$ the number of classes). To some extent, one can say that non-differentiability is our trade-off to non-parametric.

%\subsection{Exploitative optimization versus explorative optimization}
%\label{sec:exploitative-explorative}
%Given a self-supervised learning approach, we use the definition in below to determine whether it is \emph{exploitative}, or \emph{explorative}.

\begin{definition}\label{def_1}
Let $\mathbf{P}$ be a data distribution over $\R^{H \times W \times C}$. Let $\Theta=\{\theta_i\}_{i=1}^I$ be a set of parameters, where each $\theta_i$ is either network parameters (such as in a main model $f_{\theta_i} : \R^{H \times W \times C} \to \R^{K}$ or in a latent projector $f_{\theta_i} : \R^{K_{i}^1} \to \R^{K_{i}^2}$) or supportive parameters (such as codebook vectors \citep{hu2023robust} or memory queue \citep{he2020momentum}). Let $\mathcal{B} = \{\mathbf{X}_n\}_{n=1}^N$ be a batch of samples with $\mathbf{X}_n\sim \mathbf{P}$. Let $\mathcal{E}: \{\mathbf{X}_n\}_{n=1}^N \to \{\tilde{\mathbf{X}}_{n'}\}_{n'=1}^{N'}$ be the random perturbation applied to $\mathcal{B}$ (such as masking or augmentation), and let $\epsilon$ be the random perturbation applied to $\Theta$ (such as dropout). Finally, let $\Phi$ be a self-supervised learning (SSL) approach, and let $\phi\!\left(\Theta,\, \mathcal{B}\right) \in \R$ be its fitness score.

We say that $\Phi$ is \textbf{exploitative}, if there exists a map $\Omega(\phi,\,  \Theta,\,  \mathcal{B},\,  \mathcal{E})$ taking values in the parameter-update space, such that the actual parameter updates can be written as
\begin{equation}
    \delta\Theta \;=\; \Omega(\phi,\,  \Theta,\,  \mathcal{B},\,  \mathcal{E}) \;+\; \epsilon ,
\end{equation}
where both $\mathcal{E}$ and $\epsilon$ are applied directly without performance-based selection. That is, $\phi$ is never evaluated at multiple $\Theta + \epsilon$ settings or multiple $\mathcal{E}(\mathcal{B})$ settings in order to decide $\delta\Theta$.

Otherwise (if such conditions on $\Omega$, $\mathcal{E}$ and $\epsilon$ cannot be satisfied), we say that $\Phi$ is \textbf{explorative}.
\end{definition}
Intuitively, an exploitative approach determines its future by its past, whereas an explorative approach refuses to be confined by its past and embraces any change that benefits its learning goal. From this perspective, explorative SSL approaches are more general. In Appendix \ref{SSL eval}, we show that: (i) \textbf{all dominant self-supervised image clustering approaches are exploitative;} (ii) DeepDPM \citep{ronen2022deepdpm} and UNSEEN \citep{leiber2024dying}, the two baseline models in the next section, are explorative and exploitative, respectively; (iii) if we expand our scope, then generative adversarial networks (GAN) \citep{goodfellow2020generative} and diffusion models \citep{ho2020denoising, peebles2023scalable} are also exploitative, while reinforcement learning \citep{sutton1998reinforcement} is explorative. 

Now, we propose another fundamental hypothesis in this paper:
\begin{hypothesis}\label{hypothesis_1}
Let $\mathbf{P},\, \Theta,\, \mathcal{B},\,  \mathcal{E}$ and $\epsilon$ be the data distribution, set of model parameters, set of samples and perturbations defined as in above. Let $\mathcal{S}\!\left(\Theta,\, \mathcal{B},\, \mathcal{E},\, \epsilon \right) \in \R$ be a surprise score measuring, in any manner, how non-random the model's output representation is. Let $\Phi$ be an SSL approach.

If $\Phi$ is exploitative, then $\mathcal{S}\!\left(\Theta,\, \mathcal{B},\, \mathcal{E},\, \epsilon \right)$ cannot, in general, be fully optimized by $\Phi$.
\end{hypothesis}

In Section \ref{sec:exp-ablation}, we transform $\mathcal{S}(\theta)$ into two differentiable formats and train the models by gradient descent, so that both are exploitative approaches. Both attempts perform less satisfying than our converge-to-surprise scheme. Although not a proof, such results are experimental evidences supporting Hypothesis \ref{hypothesis_1}. More details regarding these two differentiable transformations are provided in Appendix \ref{IIC and Gumbel}.

From another perspective, our surprise score $\mathcal{S}(\theta)$ increases both alignment and uniformity \citep{wang2020understanding, mikulasch2026understanding} at the same time. The observed view-matching rate $\hat{q}_k$ is the alignment term. By increasing it, we pull the two views of one image into the same cluster. The marginal clustering rate $p_k$ is the uniformity term. By balancing the distribution $\{p_k\}_{k=1}^K$, we spread the clustering of different images over all $K$ output dimensions. More discussions are provided in Appendix \ref{appen_c}.

%In Appendix \ref{IIC and Gumbel}, we transform $\mathcal{S}(\theta)$ into the softmax mutual information maximization format in IIC \citep{ji2019invariant}, or replace the argmax operation in $\mathcal{S}(\theta)$ by Gumbel-softmax \citep{jang2016categorical} (while keeping $K$ unchanged). These two modifications lead to exploitative approaches. The model is trained end-to-end under these two modifications, and fails in both cases. Although not a strict proof, such results in Appendix \ref{IIC and Gumbel} are experimental evidences supporting Hypothesis \ref{hypothesis_1}.

%As mentioned in Introduction \ref{sec:intro}, strictly proving this hypothesis is beyond the scope of this paper. But more analysis is provided in Appendix \ref{IIC and Gumbel}. Since Hypothesis \ref{hypothesis_1} disables all exploitative approaches, we rely on evolution strategy to optimize $\mathcal{S}(\theta)$.

\subsection{Optimization}
\label{sec:optimization}

\paragraph{Evolution strategy.} We maximize the surprise score $\mathcal{S}(\theta)$ with the evolution strategy (ES) described in \citep{salimans2017evolution}. That is, we treat $\mathcal{S}(\theta)$ as the fitness score of a black-box optimization problem over the model's flat parameter vector $\theta \in \mathbb{R}^{D}$.

At each ES step, we sample a population of $m$ perturbed parameter vectors around the current $\theta$. To be specific, we split the population into $m/2$ mirrored pairs: For each pair, we draw $\epsilon_i \sim \mathcal{N}(0,\, \sigma^2 I_D)$, which forms two models with parameter $\theta + \epsilon_i$ and $\theta - \epsilon_i$, respectively. Here, $\mathcal{N}(0,\, \sigma^2 I_D)$ is the $D$-dim Gaussian distribution with variance $\sigma$ \citep{mackay1998introduction_to_gaussian}.

Given $N$ images $\{\mathbf{X}_n\}_{n=1}^N \sim \mathbf{P}$, we obtain the complementary masked and independently augmented views $\{\tilde{\mathbf{X}}_n^{(i)}\}_{n=1}^N$ and $\{\tilde{\mathbf{X}}_n^{(j)}\}_{n=1}^N$, which are shared across all models in all $m/2$ pairs. We implement both models in each pair to obtain the fitness scores $\mathcal{S}(\theta + \epsilon_i)$ and $\mathcal{S}(\theta - \epsilon_i)$ for $i=1,\cdots,m/2$. Then, we rank-shape \citep{wierstra2014natural} all $m$ scores into centered ranks $r_i \in [-\tfrac12, \tfrac12]$, and apply a weighted update to $\theta$:
\begin{equation}
    \theta \;\leftarrow\; (1 - \eta\lambda)\,\theta
    \;+\; \frac{\eta}{m\sigma}\,\sum\nolimits_{i=1}^{m} r_i\,\epsilon_i.
    \label{eq:es-update}
\end{equation}
Here, $\eta$ is the learning rate  and $\lambda$ is the weight decay rate. Note that each $r_i$ evolves performance-based selection (ranking), which is directly multiplied by each random perturbation $\epsilon_i$. Hence, we cannot find a map $\Omega$ in Definition \ref{def_1} to represent formula \ref{eq:es-update}. Thus, our approach is explorative.

Mirrored pairs, rank-shaping and weight-decay are commonly used tricks in deep evolution strategies \cite{wierstra2014natural, salimans2017evolution}. More discussions are in \citep{salimans2017evolution}. %In addition, we apply data augmentation \citep{perez2017effectiveness} \emph{independently} on each view of $\mathbf{X}^{(i)}$ and $\mathbf{X}^{(j)}$ from one image $\mathbf{X}$ to further stabilize the optimization process. More details are described in Section \ref{experiments}.

\paragraph{Surrogate training using surprising results.}
Evolution strategy (ES) on its own is sufficient to discover clusters with surprising view-matching results. But it gives equal credit to every output dimension $k$ contributing to $\mathcal{S}(\theta)$, including tiny, noisy micro-clusters that share similar features with larger clusters. 

So, we first apply $T_0$ epochs of pure ES optimization to warm-up, during which clusters with surprising view-matching results will emerge. Starting from epoch $T_0+1$, we apply $T_{1,a}$ epochs of gradient-descent training every $T_{1,b}$ epochs of ES optimization. They form a complete period with $T_{1,a} + T_{1,b}$ epochs. We implement multiple such periods. 

At the beginning of a gradient-descent training epoch, we identify the set of \emph{contributing positions} among the sampled $N$ original images $\{\mathbf{X}_1, \dots, \mathbf{X}_N\}$:
\begin{equation}
    \mathcal{C} \!=\! \Bigl\{\! n \!\in\! \{1,\dots,N\}\!:
    \hat y_n^{(i)} = \hat y_n^{(j)} = k
    \;\text{and}\; D(\hat q_k \,\Vert\, q_k) \ge \tau \!\Bigr\},
    \label{eq:contributing}
\end{equation}
where $\hat y_n^{(i)} = \arg\max_k [f_\theta(\mathbf{X}_n^{(i)})]_k$ is the predicted cluster of the un-augmented view $\mathbf{X}_n^{(i)}$ of the original image $\mathbf{X}_n$, $\hat y_n^{(j)}$ is obtained accordingly on the un-augmented view $\mathbf{X}_n^{(j)}$, $D(\hat q_k \,\Vert\, q_k)$ is the surprise score at cluster $k$ defined in formula \ref{surprise_score_k}, and $\tau$ is a pre-defined threshold on the per-cluster surprise score. To be specific, we regard cluster $k$ as \emph{surprising} if $D\!\left(\hat q_k \,\Vert\, q_k\right) \geq \tau$.

That says, $n$ is a contributing position when both views of $\mathbf{X}_n$ are predicted to the same cluster, and the view-matching result of the predicted cluster is surprising enough under $\mathcal{H}_0$. We obtain $\mathcal{C}$ on un-augmented views to avoid unnecessary perturbations during view-matching observation.

Then, we group the original images at contributing positions by their predicted cluster: $$\mathcal{C}_k=\{\mathbf{X}_n: n \in \mathcal{C} \ \text{and} \ \hat y_n^{(i)} = k\}.$$ 
Given the per-cluster count $n_k = |\mathcal{C}_k|$, let $M$ be the median of $\{n_k\}_{ D(\hat q_k \,\Vert\, q_k) \ge \tau}$. Then, within each $\mathcal{C}_k$, we select $\min(n_k,\, M)$ images uniformly without replacement: large clusters are sub-sampled down to $M$ images, whereas small clusters contribute all their images. Combining all selected images, we obtain a training set $\mathcal{B}$ with no single dominant cluster.

For each selected image $\mathbf{X}_n \in \mathcal{B}$, we use its predicted cluster $\hat y_n^{(i)}$ as the surrogate label, denoted as $\hat y_n$. We apply chessboard masking and independent augmentation again to obtain new $\tilde{\mathbf{X}}_n^{(i)}$ and $\tilde{\mathbf{X}}_n^{(j)}$ from $\mathbf{X}_n$. The model is trained by gradient descent via cross-entropy loss \citep{mao2023cross}:
\begin{equation}
    \mathcal{L}_\text{ft}(\theta) = \sum_{n \in \mathcal{B}} \Bigl[
        \mathrm{CE}\bigl(f_\theta(\tilde{\mathbf{X}}_n^{(i)}),\, \hat y_n\bigr)
        + \mathrm{CE}\bigl(f_\theta(\tilde{\mathbf{X}}_n^{(j)}),\, \hat y_n\bigr) \Bigr].
\end{equation}
Finally, algorithm~\ref{alg:cts} in Appendix \ref{CtS Algorithm} summarizes the full optimization procedure.

\section{Experiments}\label{experiments}

% Tighten the space before \paragraph headings within this section only.
\makeatletter
\renewcommand\paragraph{\@startsection{paragraph}{4}{\z@}%
  {2ex \@plus 0.5ex \@minus .2ex}%
  {-1em}%
  {\normalfont\normalsize\bfseries}}
\makeatother

We evaluate our converge-to-surprise scheme in non-parametric self-supervised image clustering, the strictest image clustering task in which the number of ground-truth classes is unknown in training. We report results on three simple benchmarks and compare against the leading non-parametric deep-clustering methods, DeepDPM \citep{ronen2022deepdpm} and UNSEEN \citep{leiber2024dying}.

\subsection{Experimental setup}
\label{sec:exp-setup}

\textbf{Model and data:}
Our model $f_\theta$ is a ResNet-9 \citep{he2016deep_residual_learning} whose linear head returns $K=64$ logits. A view's predicted cluster is the argmax over the $\ell_2$-normalized logits. Since the model is never told the number of classes, $K$ acts only as an upper bound, and the number of active clusters is discovered during training. The same architecture is used for all three benchmarks, each with ten ground-truth classes: \textbf{MNIST} \citep{MNIST_paper}, which contains $70{,}000$ training images of hand-written digits at a $28\times28$ resolution; \textbf{Fashion-MNIST} \citep{xiao2017fashion}, which matches MNIST in size and resolution but depicts clothing items in grayscale (and hence more challenging than MNIST); and \textbf{USPS} \citep{hull2002database}, which contains $7{,}291$ training images at a $16\times16$ resolution and a notable class imbalance. After the chessboard masking of Section~\ref{sec:masking}, the two complementary views are augmented independently --- random rotation, cropping and brightness/contrast jitter, plus a horizontal flip on Fashion-MNIST and an anisotropic zoom-out on USPS that decorrelates the aspect-ratio variation of its dominant digit-0 class. More details can be found in Appendix~\ref{appen_impl}.

\textbf{Optimization and protocol:}
We maximize $\mathcal{S}(\theta)$ with an ES outer loop of population size $m=32$, paired with a gradient-descent training inner loop on the discovered surprising clusters ($\tau=0.005$ in Eq.~\ref{eq:contributing}). MNIST and Fashion-MNIST use a two-stage schedule: $2000$ epochs of pure ES, then $4$ gradient-descent epochs every $25$ ES epochs until epoch $3000$. USPS inserts an intermediate weak stage ($2$ gradient-descent epochs every $500$ ES epochs for 4000 epochs) after 4000 pure ES epochs, followed by the same last 1000 epochs as in MNIST. This is because USPS has fewer samples, so models need more epochs to evolve. The surprise score is computed over batches of $N=3000$ images ($3650$ for USPS, chosen so that each epoch is split into two nearly equal steps, producing almost the same statistics). We report the mean $\pm$ standard deviation over ten independent runs from fresh initializations. At test time, one masked view (without augmentation) of each image is passed through the network and assigned to its argmax cluster. Training one model takes roughly 5 hours on MNIST or Fashion-MNIST and 2 hours on USPS, using a single NVIDIA H200 GPU.

\textbf{Metrics and baselines:}
We report clustering accuracy (ACC) under the Kuhn-Munkres (Hungarian) algorithm \citep{kuhn1955hungarian}, Normalized Mutual Information (NMI) \citep{kvaalseth2017normalized} and the Adjusted Rand Index (ARI) \citep{santos2009use} --- all higher-is-better, defined as in \citep{ronen2022deepdpm} --- together with the inferred number of active clusters $\hat K$. Baselines are the values reported by \citep{ronen2022deepdpm} and \citep{leiber2024dying}. Again, more details are in Appendix~\ref{appen_impl}.

\begin{table*}[t]
    \centering
    \small
    \setlength{\tabcolsep}{1.0pt}
    \caption{Non-parametric clustering results (\%, mean $\pm$ std) on MNIST, Fashion-MNIST, and USPS. For each metric (NMI, ARI, ACC) higher is better, and the best mean per column is in \textbf{bold}. Baselines are the values reported by \citep{ronen2022deepdpm} (DBSCAN \citep{ester1996density}, moVB \citep{hughes2013memoized}, DPM Sampler and DeepDPM \citep{ronen2022deepdpm}; on pretrained-autoencoder features) and \citep{leiber2024dying} (UNSEEN+DCN \citep{yang2017towards}, UNSEEN+DEC \citep{xie2016unsupervised}, UNSEEN+DKM \citep{fard2020deep}; on an autoencoder backbone). Our results are over ten independent runs from fresh initializations.}
    \label{tab:main-results}
    \begin{tabular}{l ccc ccc ccc}
        \toprule
        & \multicolumn{3}{c}{MNIST} & \multicolumn{3}{c}{Fashion-MNIST} & \multicolumn{3}{c}{USPS} \\
        \cmidrule(lr){2-4}\cmidrule(lr){5-7}\cmidrule(lr){8-10}
        Method & NMI & ARI & ACC & NMI & ARI & ACC & NMI & ARI & ACC \\
        \midrule
        DBSCAN       & $92.0_{\pm0.0}$ & $86.0_{\pm0.0}$ & $89.0_{\pm0.0}$ & $63.0_{\pm0.0}$ & $32.0_{\pm0.0}$ & $39.0_{\pm0.0}$ & $72.0_{\pm0.0}$ & $46.0_{\pm0.0}$ & $57.0_{\pm0.0}$ \\
        moVB        & $93.0_{\pm0.0}$ & $94.0_{\pm0.0}$ & $97.0_{\pm0.0}$ & $66.0_{\pm2.0}$ & $47.0_{\pm3.0}$ & $55.0_{\pm3.0}$ & $87.0_{\pm2.0}$ & $86.0_{\pm4.0}$ & $90.0_{\pm4.0}$ \\
        DPM Sampler                            & $92.0_{\pm1.0}$ & $91.0_{\pm4.0}$ & $93.0_{\pm5.0}$ & $67.0_{\pm1.0}$ & $49.0_{\pm2.0}$ & $59.0_{\pm3.0}$ & $87.0_{\pm1.0}$ & $82.0_{\pm2.0}$ & $83.0_{\pm3.0}$ \\
        DeepDPM       & $94.0_{\pm0.0}$ & $95.0_{\pm0.0}$ & $\mathbf{98.0_{\pm0.0}}$ & $\mathbf{68.0_{\pm1.0}}$ & $51.0_{\pm2.0}$ & $62.0_{\pm3.0}$ & $88.0_{\pm0.0}$ & $86.0_{\pm1.0}$ & $89.0_{\pm2.0}$ \\
        \midrule
        UNSEEN+DCN     & $87.8_{\pm1.9}$ & $83.7_{\pm4.5}$ & $86.0_{\pm5.4}$ & $63.7_{\pm1.0}$ & $46.9_{\pm1.5}$ & $57.3_{\pm2.4}$ & $76.5_{\pm1.4}$ & $69.8_{\pm4.9}$ & $77.5_{\pm5.6}$ \\
        UNSEEN+DEC & $80.4_{\pm1.8}$ & $70.9_{\pm3.5}$ & $71.0_{\pm4.4}$ & $58.6_{\pm2.6}$ & $42.3_{\pm3.4}$ & $54.4_{\pm3.8}$ & $80.8_{\pm0.9}$ & $75.9_{\pm2.4}$ & $81.3_{\pm2.4}$ \\
        UNSEEN+DKM      & $84.2_{\pm2.6}$ & $78.5_{\pm5.2}$ & $83.0_{\pm5.5}$ & $62.2_{\pm1.0}$ & $43.9_{\pm1.4}$ & $53.1_{\pm2.0}$ & $73.3_{\pm5.3}$ & $62.6_{\pm10.6}$ & $71.5_{\pm9.4}$ \\
        \midrule
        Ours                      & $\mathbf{95.6_{\pm0.7}}$ & $\mathbf{96.1_{\pm1.0}}$ & $97.9_{\pm1.3}$ & $64.0_{\pm1.7}$ & $\mathbf{51.9_{\pm2.2}}$ & $\mathbf{64.1_{\pm1.8}}$ & $\mathbf{89.4_{\pm1.7}}$ & $\mathbf{89.1_{\pm3.0}}$ & $\mathbf{93.4_{\pm3.1}}$ \\
        \bottomrule
    \end{tabular}
\end{table*}

\subsection{Results}
\label{sec:exp-results}

Table~\ref{tab:main-results} reports the comparison. Our framework achieves state-of-the-art performances on USPS across all three metrics. On MNIST, our model leads in NMI and ARI while almost matching DeepDPM in ACC ($97.9\%$ against $98.0\%$, though with a larger variance). On Fashion-MNIST, we obtain the best ACC and ARI scores, while remaining competitive in NMI. The largest improvement occurs on USPS, where our model improves ACC from DeepDPM's $89\%$ to $93.4\%$, and lifts NMI/ARI by roughly $1.5\%$ and $3\%$, respectively. On Fashion-MNIST, we reach $64.1\%$ ACC and $51.9\%$ ARI, both above the strongest baselines, whereas our NMI of $64.0\%$ is second to DeepDPM's $68\%$. In addition, our model is trained from scratch, rather than from pre-trained features.

\begin{table}[t]
    \centering
    \small
    \setlength{\tabcolsep}{5pt}
    \caption{Inferred number of clusters $\hat K$ (mean $\pm$ std) on the three benchmarks, with the ground-truth value to be $K^\star=10$. Baseline values are as reported by \citep{ronen2022deepdpm}. With no prior knowledge, our discovered number of active clusters are close to the ground truth.}
    \label{tab:inferred-k}
    \begin{tabular}{l ccc}
        \toprule
        Method & MNIST & Fashion-MNIST & USPS \\
        \midrule
        DBSCAN \citep{ester1996density}  & $9.0_{\pm0.0}$  & $4.0_{\pm0.0}$  & $6.0_{\pm0.0}$  \\
        DPM Sampler                       & $11.3_{\pm0.8}$ & $12.4_{\pm1.0}$ & $8.5_{\pm0.9}$  \\
        moVB \citep{hughes2013memoized}   & $14.0_{\pm1.0}$ & $16.9_{\pm2.3}$ & $11.2_{\pm1.1}$ \\
        DeepDPM \citep{ronen2022deepdpm}  & $10.0_{\pm0.0}$ & $10.2_{\pm0.8}$ & $9.2_{\pm0.4}$  \\
        \midrule
        Ours                              & $10.1_{\pm0.3}$ & $11.2_{\pm1.0}$ & $10.2_{\pm0.8}$ \\
        \bottomrule
    \end{tabular}
\end{table}

Crucially, these results are obtained without telling the model the number of true classes. As shown in Table~\ref{tab:inferred-k}, the number of active clusters our method discovers stays close to the ground-truth value of ten: $10.1_{\pm0.3}$ on MNIST, $11.2_{\pm1.0}$ on Fashion-MNIST, and $10.2_{\pm0.8}$ on USPS. These results are comparable to the values inferred by DeepDPM ($10.0_{\pm0.0}$, $10.2_{\pm0.8}$ and $9.2_{\pm0.4}$, respectively) and largely beat the values by other baselines. The remaining $K-\hat K$ output dimensions naturally die out during our training. Appendix \ref{appen_num_clusters} shows how the number of active clusters changes across epochs.

In addition, DeepDPM is explorative, but applies pre-trained autoencoder features. The autoencoder in UNSEEN is trained from scratch, but its `dying cluster' mechanism is exploitative (further discussions are provided in Appendix \ref{ssl_eval_baselines}). To the edge of our knowledge, our model is the first explorative SSL approach being trained from scratch for self-supervised image clustering. 

In summary, simply by maximizing the surprise score --- without pre-trained models or prior knowledge --- a deep network can naturally discover semantically meaningful hard representation from simple digit or garment images. Visualizations of the clustered images are exhibited in appendix \ref{appen_visualization}.

Models trained by our scheme perform less satisfying on more complicated datasets such as CIFAR10 \citep{krizhevsky2009learning}. More details are provided in Appendix \ref{fail_CIFAR10}.

\subsection{Ablation studies}
\label{sec:exp-ablation}

In this sub-section, we first replace the explorative optimizer (converge-to-surprise scheme) by two exploitative ones. Then, we ablate three major components of our approach: the independent two-view augmentation, the inner-loop gradient-descent training, and the anisotropic (zoom-out) augmentation used only on the USPS dataset. Table~\ref{tab:ablation} reports the effect.

\textbf{Exploitative approaches do not match evolution strategy.}
The surprise score can be made differentiable and then maximized by gradient descent. We replace the argmax of Eq.~\ref{eq:argmax_logit} by either a plain softmax head, or a Gumbel softmax \citep{marques2015distribution} head. That is,
\begin{align}\label{eq:app_plain_assign_main}
    \boldsymbol{\pi}_n &= \mathrm{softmax}\!\left(\mathbf{y}_n^{(i)} / \tau'\right),
    \qquad
    \boldsymbol{\rho}_n = \mathrm{softmax}\!\left(\mathbf{y}_n^{(j)} / \tau'\right); \ \ \ \textrm{or}\\
    \boldsymbol{\pi}_n &= \mathrm{softmax}\!\left(\mathbf{y}_n^{(i)} / \tau' + \mathbf{g}_n^{(i)}\right),
    \qquad
    \boldsymbol{\rho}_n = \mathrm{softmax}\!\left(\mathbf{y}_n^{(j)} / \tau' + \mathbf{g}_n^{(j)}\right).
\end{align}
where $\mathbf{g}_n^{(i)}, \mathbf{g}_n^{(j)}$ are drawn i.i.d.\ from a Gumbel variable between 0 and 1 \citep{marques2015distribution}, independently for each image, each cluster and each of the two views, and re-drawn at every step. Also, $\tau'$ is the temperature in the softmax operation, which should not be confused with the surprising cluster threshold $\tau$ used in formula \ref{eq:contributing}.

Then, every count of Section~\ref{sec:surprise} becomes a soft expectation,
\begin{equation}\label{eq:app_plain_counts_main}
    \hat q_k = \frac{1}{N}\sum_{n=1}^{N} \pi_{n,k}\,\rho_{n,k},
    \qquad
    p_k = \frac{1}{2N}\sum_{n=1}^{N}\left(\pi_{n,k} + \rho_{n,k}\right),
    \qquad
    q_k = p_k^{\,2}.
\end{equation}
Also, the over-matching indicator of Eq.~\ref{eq:surprise} can be replaced by a smooth gate using the sigmoid function \citep{kyurkchiev2015_sigmoid}, which produces the differentiable objective to be used as the final loss:
\begin{equation}\label{eq:app_plain_score_main}
    \mathcal{S}_{\mathrm{soft}}(\theta;\tau,\varepsilon)
    = \sum\nolimits_{k} \sigma\!\left(\frac{\hat q_k - q_k}{\varepsilon}\right) D\!\left(\hat q_k \,\Vert\, q_k\right).
\end{equation}
Both approaches lead to differentiable learning targets, and hence are exploitative by Definition \ref{def_1}. We train both models using gradient descent, with every other setting exactly the same as the converge-to-surprise scheme. Table~\ref{tab:ablation} shows that although both approaches produce non-trivial results, our converge-to-surprise scheme still surpasses them in eight out of nine columns. Although not a strict proof, such experimental results are highly supportive to Hypothesis \ref{hypothesis_1}, saying that exploitative approaches cannot fully optimize a surprise score. Once again, more details regarding these two differentiable transformations can be found in Appendix \ref{IIC and Gumbel}.

\begin{table*}[t]
    \centering
    \small
    \setlength{\tabcolsep}{0.5pt}
    \caption{Ablation studies (\%, mean $\pm$ std; the full model over ten runs, each ablation over five runs). The first two rows replace the evolution strategy by gradient descent on a differentiable surrogate of the surprise score; each remaining row removes one component from the full model (GDT is short for gradient-descent training). The anisotropic zoom-out augmentation is USPS-specific (`--').}
    \label{tab:ablation}
    \begin{tabular}{l ccc ccc ccc}
        \toprule
        & \multicolumn{3}{c}{MNIST} & \multicolumn{3}{c}{Fashion-MNIST} & \multicolumn{3}{c}{USPS} \\
        \cmidrule(lr){2-4}\cmidrule(lr){5-7}\cmidrule(lr){8-10}
        Setting & NMI & ARI & ACC & NMI & ARI & ACC & NMI & ARI & ACC \\
        \midrule
        conv.-to-surprise             & $\mathbf{95.6_{\pm0.7}}$ & $\mathbf{96.1_{\pm1.0}}$ & $\mathbf{97.9_{\pm1.3}}$ & $64.0_{\pm1.7}$ & $\mathbf{51.9_{\pm2.2}}$ & $\mathbf{64.1_{\pm1.8}}$ & $\mathbf{89.4_{\pm1.7}}$ & $\mathbf{89.1_{\pm3.0}}$ & $\mathbf{93.4_{\pm3.1}}$ \\
        plain-softmax  & $91.3_{\pm3.7}$ & $81.0_{\pm9.0}$ & $82.8_{\pm8.4}$ & $60.6_{\pm3.5}$ & $46.5_{\pm4.4}$ & $58.5_{\pm6.2}$ & $80.4_{\pm4.3}$ & $70.6_{\pm9.0}$ & $72.1_{\pm8.1}$ \\
        Gumbel-softmax & $93.3_{\pm2.0}$ & $87.6_{\pm6.0}$ & $88.2_{\pm7.4}$ & $\mathbf{64.3_{\pm2.3}}$ & $50.2_{\pm3.9}$ & $60.5_{\pm4.2}$ & $84.4_{\pm5.3}$ & $77.0_{\pm9.7}$ & $82.7_{\pm8.3}$ \\
        no augmentation        & $49.1_{\pm2.5}$ & $25.1_{\pm1.9}$ & $30.1_{\pm1.5}$ & $49.8_{\pm0.6}$ & $23.5_{\pm1.1}$ & $26.9_{\pm1.7}$ & $65.3_{\pm3.0}$ & $47.7_{\pm5.1}$ & $55.2_{\pm5.9}$ \\
        no inner-loop GDT       & $77.4_{\pm3.0}$ & $74.3_{\pm5.3}$ & $80.9_{\pm6.1}$ & $58.1_{\pm2.1}$ & $44.9_{\pm2.2}$ & $56.6_{\pm1.5}$ & $66.2_{\pm5.1}$ & $51.7_{\pm9.2}$ & $59.3_{\pm9.3}$ \\
        no anisotropic aug.\   & -- & -- & -- & -- & -- & -- & $80.4_{\pm2.8}$ & $73.0_{\pm4.8}$ & $83.4_{\pm3.3}$ \\
        \bottomrule
    \end{tabular}
\end{table*}

\textbf{Augmentation is essential.}
Without augmentation, evolution strategy (ES) maximizes the surprise score by latching onto low-level features that the two views happen to share --- stroke thickness, brightness, and overall shape --- rather than the semantic class. Augmenting the two views independently breaks this shortcut: the two views now differ in brightness, contrast, scale (cropping) and orientation (rotation). So, only augmentation-invariant features can make the two views agree, which are more likely the true underlying topology of the digits or items.

\textbf{Inner-loop gradient-descent training is important.}
After several thousand epochs of pure ES, the model consistently produces more clusters than the number of ground-truth classes. Gradient-descent training then consolidates discovered clusters: if two clusters hold similar image views, the larger one will provide stronger pseudo-label signals, which finally pulls across the views from the smaller one. Augmenting both views again before each training step further reinforces the clusters discovered by ES. The two-view agreement rate rises from $80$--$85\%$ after pure ES to nearly $99\%$. %However, without evolution epochs, there is no discovered surprising cluster for gradient-descent training to consolidate. The less satisfying performances from the exploitative optimizations show the importance of evolution strategy.

\textbf{Anisotropic augmentation is necessary for USPS.}
The USPS dataset is class-imbalanced, and its digit-zero samples vary widely in aspect ratio (tall-and-thin versus short-and-wide). The single-axis-zoom-out randomly stretches or compresses each view along one axis, so the two views of an image differ in width and height. Aspect ratio is then no longer a salient feature shared by the two views. This forces the network to explore more essential, topology-related structures of digits.

% Restore the default \paragraph spacing after the Experiments section.
\makeatletter
\renewcommand\paragraph{\@startsection{paragraph}{4}{\z@}%
  {3.25ex \@plus 1ex \@minus .2ex}%
  {-1em}%
  {\normalfont\normalsize\bfseries}}
\makeatother

\section{Conclusion}
\label{sec:conclusion}

In this paper, we introduce converge-to-surprise, a self-supervised learning framework aiming at discovering non-randomness from images. Our null hypothesis states that pixels are i.i.d.\ random noise. Then, we build two complementary masked views from one image, and define a surprise score to measure how strongly a deep learning model's view-clustering results reject the null hypothesis. Assuming that a surprise score cannot be fully optimized by exploitative SSL approaches, we optimize our model combining evolution strategy with gradient descent, leading to an explorative approach. Experiments show that our model trained from scratch attains state-of-the-art performances on simple image benchmarks like MNIST, Fashion-MNIST and USPS, under the strictest non-parametric self-supervised image clustering setting. To the edge of our knowledge, our model is the first explorative SSL approach being trained from scratch for self-supervised image clustering.

\bibliographystyle{iclr2027_conference}
\bibliography{references}

\appendix

\section{Evaluating existing self-supervised learning approaches for exploitative and explorative}
\label{SSL eval}

In this section, we apply Definition \ref{def_1} to the dominant families of self-supervised learning. In each case, we explicitly define $\Omega$. For every gradient-based approach below, $\phi$ is the training loss up to a sign, and $\Omega$ is one gradient step on it.

Note that in Definition \ref{def_1}, $\Theta$ denotes the complete current state of one SSL approach --- including network parameters, codebook vectors, memory queue, optimizer momentum, etc. This is also why $\Theta = \{\theta_i\}_{i=1}^{I}$ is defined as a set of parameters rather than a single vector. With that reading, none of the approaches below has parameters not covered by $\Theta$.

\subsection{Self-distillation}
\label{ssl_eval_selfdistill}

We first evaluate the self-distillation family, which includes BYOL \citep{grill2020bootstrap} and the DINO family: DINO \citep{caron2021emerging}, DINOv2 \citep{oquab2023dinov2} and DINOv3 \citep{simeoni2025dinov3}. Here, the state consists of two networks, a student and a teacher:
\begin{equation}
    \Theta \;=\; \left\{\theta_s,\; \theta_t,\; c\right\},
\end{equation}
where $\theta_s$ and $\theta_t$ are the student and teacher network parameters, and $c$ is the centering vector (not a constant) used by the DINO family (absent in BYOL, which uses a predictor head instead).

\textbf{The input perturbation.} From each image, several views are extracted at different scales --- two global crops and a number of smaller local crops, together with photometric jitter. This is precisely the random perturbation $\mathcal{E}$ in Definition \ref{def_1} being applied to the batch $\mathcal{B}$, and it is the reason that $\mathcal{E}$ is allowed to change the number of samples, $\mathcal{E} : \{\mathbf{X}_n\}_{n=1}^{N} \to \{\tilde{\mathbf{X}}_{n'}\}_{n'=1}^{N'}$. We use $\mathcal{V} = \mathcal{E}(\mathcal{B})$ to denote the resulting set of views and use $\mathcal{V}_g \subset \mathcal{V}$ to denote the global ones.

\textbf{The training loss.} The student is trained to match the teacher's output. In DINO, both networks end in a $K$-dimensional head, which is turned into a distribution by a softmax. The softmax is further sharpened by temperatures $\tau_s, \tau_t > 0$, with the teacher's output additionally centered by $c$:
\begin{equation}\label{eq:dino_heads}
    \mathbf{P}_s(\tilde{\mathbf{X}}) = \mathrm{softmax}\!\left(\frac{f_{\theta_s}(\tilde{\mathbf{X}})}{\tau_s}\right),
    \qquad
    \mathbf{P}_t(\tilde{\mathbf{X}}) = \mathrm{softmax}\!\left(\frac{f_{\theta_t}(\tilde{\mathbf{X}}) - c}{\tau_t}\right).
\end{equation}
The loss is the cross-entropy between the teacher's distribution on a global view and the student's distribution on every other view:
\begin{equation}\label{eq:dino_loss}
    \mathcal{L}\!\left(\theta_s,\, \theta_t,\, c,\, \mathcal{E}(\mathcal{B})\right)
    \;=\; -\sum_{\tilde{\mathbf{X}} \in \mathcal{V}_g} \;\;
    \sum_{\substack{\tilde{\mathbf{X}}' \in \mathcal{V} \\ \tilde{\mathbf{X}}' \neq \tilde{\mathbf{X}}}} \;\;
    \sum_{k=1}^{K}
    \left[\mathbf{P}_t(\tilde{\mathbf{X}})\right]_k \,
    \log \left[\mathbf{P}_s(\tilde{\mathbf{X}}')\right]_k .
\end{equation}
BYOL differs in the form of the target but not in its structure: a predictor $q_{\theta_s}$ is applied on the student branch and the loss is the squared distance between $\ell_2$-normalized projections, $\bigl\Vert \overline{q_{\theta_s}(z_s)} - \overline{z_t} \bigr\Vert_2^{2}$, with $z_s$ and $z_t$ the student and teacher embeddings of two views of the same image. In both cases, the loss is a deterministic function of $\theta_s$, $\theta_t$, $c$, and the views $\mathcal{E}(\mathcal{B})$ --- and nothing else.

\textbf{The updates.} The student takes one gradient step. The teacher is not trained at all, but is an Exponential Moving Average (EMA) \citep{morales2024exponential} of the student, with momentum $\lambda \in (0,1)$. The centering vector is an EMA of the teacher's own output over the global views, with momentum $\mu \in (0,1)$:
\begin{align}
    \delta\theta_s &= -\eta\, \nabla_{\theta_s} \mathcal{L}\!\left(\theta_s;\, \theta_t,\, c,\, \mathcal{E}(\mathcal{B})\right),
    \label{eq:dino_student}\\[2pt]
    \delta\theta_t &= \left(1-\lambda\right)\left(\theta_s - \theta_t\right),
    \label{eq:dino_teacher}\\[2pt]
    \delta c &= \left(1-\mu\right)\left(\frac{1}{|\mathcal{V}_g|}\sum\nolimits_{\tilde{\mathbf{X}} \in \mathcal{V}_g} f_{\theta_t}(\tilde{\mathbf{X}}) \;-\; c\right).
    \label{eq:dino_center}
\end{align}
Reading Eq.~\ref{eq:dino_student}--\ref{eq:dino_center}, every component of $\delta\Theta$ is a function of the current state $\Theta = \{\theta_s, \theta_t, c\}$, the batch $\mathcal{B}$, and the perturbation $\mathcal{E}$. Also, it is easy to see that the fitness score in Definition \ref{def_1} is just the final loss, $\phi=\mathcal{L}\!\left(\theta_s;\, \theta_t,\, c,\, \mathcal{E}(\mathcal{B})\right)$. Collecting them, we have
\begin{equation}\label{eq:dino_omega}
    \Omega\!\left(\phi,\, \Theta,\, \mathcal{B},\, \mathcal{E}\right)
    \;=\; \Bigl(\, \delta\theta_s,\;\; \delta\theta_t,\;\; \delta c \,\Bigr),
    \qquad
    \delta\Theta \;=\; \Omega\!\left(\phi,\, \Theta,\, \mathcal{B},\, \mathcal{E}\right) + \epsilon ,
\end{equation}
where $\epsilon$ collects the parameter-level noise such as dropout, and is applied directly. Depending on different training settings of different approaches, dropout may or may not be actually applied. Without loss of generality, we always keep the term $\epsilon$ in this section. 

The map $\Omega$ required by Definition \ref{def_1} therefore exists, and the loss is evaluated once --- at the single setting $\left(\theta_s, \theta_t, c\right)$ and on the single view set $\mathcal{E}(\mathcal{B})$. No candidate teacher, candidate student or candidate view set is ever scored and then selected against another. Hence, \textbf{self-distillation is exploitative}.

It is worth being explicit about why the teacher does not make the approach explorative. The teacher is a second network, and one might read Eq.~\ref{eq:dino_teacher} as a form of selection between two candidates. It is not: Eq.~\ref{eq:dino_teacher} is a fixed formula being applied unconditionally, with a constant decay factor $\lambda$. The teacher is never compared against the student on $\phi$ in order to keep the better one. %This is exactly the distinction Definition \ref{def_1} draws --- randomness and perturbations are allowed, but it is the performance-based selection that makes an approach explorative.

\subsection{Contrastive learning}
\label{ssl_eval_contrastive}

We now turn to contrastive learning, taking SimCLR \citep{chen2020simple} as the example. Here, we have $\Theta = \{\theta\}$, with $\theta$ to be a single encoder combined with a projection head.

\textbf{The input perturbation.} Each image in the batch is augmented into two views. That is, $\mathcal{E} : \{\mathbf{X}_n\}_{n=1}^{N} \to \{\tilde{\mathbf{X}}_{n'}\}_{n'=1}^{2N}$. The two views of image $n$ form a positive pair and all other views act as negatives. Again, this is exactly the perturbation described in Definition \ref{def_1}.

\textbf{The training loss.} Each view is mapped to a projection $\mathbf{z}_{n'} = g_\theta\!\left(f_\theta(\tilde{\mathbf{X}}_{n'})\right)$, and views in each pair are compared by cosine similarity, $\mathrm{sim}(\mathbf{u}, \mathbf{v}) = \mathbf{u}^{\!\top}\mathbf{v} / \left(\Vert\mathbf{u}\Vert \Vert\mathbf{v}\Vert\right)$. With temperature $\tau > 0$, the loss for a positive pair $(a,b)$ and the total loss are respectively
\begin{equation}\label{eq:simclr_loss}
    \ell_{a,b} \;=\; -\log
    \frac{\exp\!\left(\mathrm{sim}(\mathbf{z}_a, \mathbf{z}_b)/\tau\right)}
         {\sum_{m \neq a} \exp\!\left(\mathrm{sim}(\mathbf{z}_a, \mathbf{z}_m)/\tau\right)},
    \qquad
    \mathcal{L}\!\left(\theta;\, \mathcal{E}(\mathcal{B})\right) = \frac{1}{2N}\sum_{(a,b)} \left(\ell_{a,b} + \ell_{b,a}\right).
\end{equation}
Here, the numerator in $\ell_{a,b}$ pulls the two views of one image together and the denominator pushes every other view away. In $\mathcal{L}\!\left(\theta;\, \mathcal{E}(\mathcal{B})\right)$, the sum runs over the $N$ positive pairs.

\textbf{The update.} Every projection $\mathbf{z}_{n'}$ in Eq.~\ref{eq:simclr_loss} is determined by the current parameters $\theta$ and the current views $\mathcal{E}(\mathcal{B})$. The loss is a function of the pairwise similarities among those projections, so it too is determined by $\theta$, $\mathcal{B}$ and $\mathcal{E}$. Again, the fitness score in Definition \ref{def_1} is the final loss, $\phi=\mathcal{L}\!\left(\theta;\, \mathcal{E}(\mathcal{B})\right)$. This will be true throughout this appendix section, unless with specify otherwise.

Finally, the update is a single gradient step,
\begin{equation}\label{eq:simclr_omega}
    \Omega\!\left(\phi,\, \Theta,\, \mathcal{B},\, \mathcal{E}\right) = -\eta\, \nabla_\theta \mathcal{L}\!\left(\theta;\, \mathcal{E}(\mathcal{B})\right),
    \qquad
    \delta\Theta = \Omega\!\left(\phi,\, \Theta,\, \mathcal{B},\, \mathcal{E}\right) + \epsilon,
\end{equation}
where $\epsilon$ is any potential perturbation. Thus, conditions in Definition \ref{def_1} are satisfied again, with $\mathcal{E}$ and $\epsilon$ being applied directly without performance-based selection. Hence, \textbf{contrastive learning is exploitative}. MoCo \citep{he2020momentum} adds a momentum encoder and a queue of stored negatives, which simply enlarges the parameter set $\Theta$ without introducing any performance-based selection.

%Note that the denominator of Eq.~\ref{eq:simclr_loss} weights negatives by their similarity, and hard-negative mining makes that ranking explicit. But this ranking happens \emph{among the samples inside one} $\mathcal{E}(\mathcal{B})$; the approach never scores two different perturbed batches $\mathcal{E}_1(\mathcal{B})$ and $\mathcal{E}_2(\mathcal{B})$ and keep the better one. Definition \ref{def_1} is phrased in terms of $\phi$ being evaluated at \emph{multiple} $\mathcal{E}(\mathcal{B})$ settings precisely so that sample-level ranking of this kind stays on the exploitative side.

\subsection{Latent-space prediction}
\label{ssl_eval_jepa}

We now evaluate the joint-embedding predictive architecture (JEPA) family \citep{lecun2022path}, taking I-JEPA \citep{assran2023self} as the example; V-JEPA \citep{bardes2024revisiting} differs only in that the blocks are drawn from a video clip rather than from a single image. Unlike the two families above, no photometric view augmentation is required here: $\mathcal{E}$ is masking alone. 

The parameter set consists of three networks,
\begin{equation}
    \Theta \;=\; \left\{\theta_c,\; \theta_t,\; \theta_p\right\},
\end{equation}
where $\theta_c$ is the context encoder, $\theta_t$ is the target encoder, and $\theta_p$ is the predictor. Here, $\theta_p$ maps a context representation --- conditioned on the position of a queried block --- to the predicted representation of that block.

\textbf{The input perturbation.} Each image $\mathbf{X}_n$ is cut into a grid of patches. From that grid, $M$ target blocks $B_{n,1}, \dots, B_{n,M}$ are sampled with random scale and aspect ratio. Then, one context block $A_n$ is sampled and has every patch overlapping any target block removed. Writing $\mathcal{E}$ for this block sampling, $\mathcal{E} : \{\mathbf{X}_n\}_{n=1}^{N} \to \left\{\left(A_n,\, B_{n,1}, \dots, B_{n,M}\right)\right\}_{n=1}^{N}$, we recover exactly the perturbation in Definition \ref{def_1}. We use $\mathbf{X}_n[A_n]$ to represent the block in its corresponding image.

\textbf{The training loss.} The target encoder is applied to the full image and produces one representation per patch, where the target for patch $j$ is $\left[f_{\theta_t}(\mathbf{X}_n)\right]_j$. The predictor is applied to the context representation and asked to reproduce that target, in latent space, for every patch of every target block:
\begin{equation}\label{eq:jepa_loss}
    \mathcal{L}\!\left(\theta_c, \theta_p;\, \theta_t,\, \mathcal{E}(\mathcal{B})\right)
    \;=\; \frac{1}{N M} \sum_{n=1}^{N} \sum_{m=1}^{M} \; \sum_{j \in B_{n,m}}
    \left\Vert\, g_{\theta_p}\!\left(f_{\theta_c}\!\left(\mathbf{X}_n[A_n]\right),\, j\right)
    \;-\; \left[f_{\theta_t}(\mathbf{X}_n)\right]_j \,\right\Vert_2^{2},
\end{equation}
with a stop-gradient on the target branch. Only the position index $j$ is passed to the predictor, whereas the content of patch $j$ is never passed to the predictor. Thus, the loss is a deterministic function of $\theta_c$, $\theta_p$, $\theta_t$ and $\mathcal{E}(\mathcal{B})$, and of nothing else. 

\textbf{The updates.} The context encoder and the predictor take one gradient step. The target encoder is an EMA of the context encoder with momentum $\lambda \in (0,1)$:
\begin{equation}\label{eq:jepa_updates}
    \delta\theta_c = -\eta\, \nabla_{\theta_c} \mathcal{L}, \qquad
    \delta\theta_p = -\eta\, \nabla_{\theta_p} \mathcal{L}, \qquad
    \delta\theta_t = \left(1-\lambda\right)\left(\theta_c - \theta_t\right).
\end{equation}
Collecting the three components, we have
\begin{equation}\label{eq:jepa_omega}
    \Omega\!\left(\phi,\, \Theta,\, \mathcal{B},\, \mathcal{E}\right)
    \;=\; \Bigl(\, \delta\theta_c,\;\; \delta\theta_p,\;\; \delta\theta_t \,\Bigr),
    \qquad
    \delta\Theta \;=\; \Omega\!\left(\phi,\, \Theta,\, \mathcal{B},\, \mathcal{E}\right) + \epsilon .
\end{equation}
The map $\Omega$ therefore exists, and $\phi$ is evaluated once --- at the single setting $\left(\theta_c, \theta_t, \theta_p\right)$ and on the single sample batch $\mathcal{E}(\mathcal{B})$. As in Appendix \ref{ssl_eval_selfdistill}, the EMA target encoder never goes through performance-based selection. Hence, \textbf{latent-space prediction is exploitative}.

%The objection worth answering here is a different one: \emph{multiplicity}. This family draws $M$ target blocks per image (and DINO, in Appendix \ref{ssl_eval_selfdistill}, draws a whole set of global and local crops), so several perturbations of the same image are present in a single step. That is not selection. Definition \ref{def_1} asks whether $\phi$ is evaluated at multiple $\mathcal{E}(\mathcal{B})$ settings \emph{in order to decide} $\delta\Theta$ --- that is, whether some candidate is kept and the others discarded on the strength of the comparison. Eq.~\ref{eq:jepa_loss} does the opposite: every one of the $M$ draws is retained, with the same weight $1/M$, and the loss incurred on block $m$ never decides whether block $m'$ is used. Averaging over draws is a Monte-Carlo estimate of an expectation over $\mathcal{E}$; it is a way of \emph{reducing} the variance contributed by the perturbation, whereas selection deliberately exploits that variance by keeping the draws that score well. The same reading applies to the energy-based presentation of JEPA \citep{lecun2022path}: the energy is descended by gradient, and collapse is prevented architecturally --- by the asymmetry between the two branches, the stop-gradient and the EMA --- rather than by comparing candidate parameter settings on $\phi$.

\subsection{Deep clustering with pseudo-labels}
\label{ssl_eval_deepcluster}

We now turn to deep clustering approaches, taking DeepCluster \citep{caron2018deep} as the example. This is the family in which one would most expect Definition \ref{def_1} to break, because its central step, $k$-means clustering, is neither a gradient step nor a differentiable operation. 

The parameter set is
\begin{equation}
    \Theta \;=\; \left\{\theta,\; W,\; C\right\},
\end{equation}
where $\theta$ is the backbone model parameters, $W$ is a linear classification head over $K$ pseudo-classes, and $C = \left[\mathbf{c}_1, \dots, \mathbf{c}_K\right]$ is the centroid matrix \citep{caron2018deep}. We can see that $C$ is the supportive parameters in Definition \ref{def_1}, in the same role as a codebook.

\textbf{The input perturbation.} Standard augmentation (random resized crop, flip, photometric jitter) is applied on each image in the batch, $\mathcal{E} : \{\mathbf{X}_n\}_{n=1}^{N} \to \{\tilde{\mathbf{X}}_{n}\}_{n=1}^{N}$. Two details of DeepCluster are worth noting, because both are sometimes mistaken for selection: (i) images are drawn uniformly over the current pseudo-classes rather than uniformly over the dataset; and (ii) the assignment process described below runs on un-augmented (or mildly augmented) images. Neither involves comparison among candidates.

\textbf{The assignment step.} Periodically (once per epoch in DeepCluster), features $\mathbf{z}_n = f_\theta(\mathbf{X}_n)$ are extracted, reduced by principle component analysis (PCA) \citep{abdi2010principal}, and then $\ell_2$-normalized, denoted as $\tilde{\mathbf{z}}_n$. After that, the processed representations $\{\tilde{\mathbf{z}}_n\}_{n=1}^N$ are clustered by Lloyd's algorithm \citep{lloyd1982least}:
\begin{equation}\label{eq:dc_kmeans}
    \left(C^{+},\; \left\{y_n\right\}_{n=1}^{N}\right)
    \;\in\; \operatorname*{arg\,min}_{C,\, \left\{y_n\right\}} \;
    \sum_{n=1}^{N} \left\Vert \tilde{\mathbf{z}}_n - \mathbf{c}_{y_n} \right\Vert_2^{2},
    \qquad y_n \in \left\{1, \dots, K\right\},
\end{equation}
which yields the pseudo-label $y_n$ of each image. Because cluster indices are permuted arbitrarily from one round to the next, the head $W$ is re-initialized whenever Eq.~\ref{eq:dc_kmeans} is recomputed.

\textbf{The training step and the updates.} With the pseudo-labels being fixed, the network is trained by multinomial logistic regression on the augmented images,
\begin{equation}\label{eq:dc_loss}
    \mathcal{L}\!\left(\theta,\, W;\, \left\{y_n\right\},\, \mathcal{E}(\mathcal{B})\right)
    \;=\; -\frac{1}{N}\sum_{n=1}^{N} \log
    \left[\mathrm{softmax}\!\left(W f_\theta(\tilde{\mathbf{X}}_n)\right)\right]_{y_n}.
\end{equation}
Thus, the complete update is
\begin{equation}\label{eq:dc_updates}
    \delta\theta = -\eta\, \nabla_{\theta} \mathcal{L}, \qquad
    \delta W = -\eta\, \nabla_{W} \mathcal{L} \;\;\text{(or a re-initialization)}, \qquad
    \delta C = C^{+} - C .
\end{equation}
Collecting these components, we have 
\begin{equation}
\Omega\!\left(\phi,\, \Theta,\, \mathcal{B}, \,\mathcal{E}\right) = \left(\delta\theta,\, \delta W,\, \delta C\right), \qquad \delta\Theta = \Omega\!\left(\phi,\, \Theta,\, \mathcal{B}, \,\mathcal{E}\right) + \epsilon.
\end{equation}
Every component of Eq.~\ref{eq:dc_updates} is computed from the current parameter set $\Theta = \{\theta, W, C\}$, the batch $\mathcal{B}$ and the perturbation $\mathcal{E}$. To be specific, the pseudo-labels of Eq.~\ref{eq:dc_kmeans} are a functional of the current output features, hence of $\theta$ and $\mathcal{B}$. The loss of Eq.~\ref{eq:dc_loss} is evaluated once without performance-based selection or comparison. Hence, \textbf{deep clustering is exploitative}. 

SwAV \citep{caron2020unsupervised} replaces the offline $k$-means by an online Sinkhorn--Knopp \citep{cuturi2013sinkhorn} balanced assignment over a set of prototypes that are themselves trained by gradient descent. The Sinkhorn step is a deterministic function of the current scores, so the verdict is unchanged.

\subsection{Masked autoencoders}
\label{ssl_eval_mae}

Now, we evaluate masked autoencoders, taking MAE \citep{he2022masked} as the example. The parameter set is
\begin{equation}
    \Theta \;=\; \left\{\theta_e,\; \theta_d,\; \mathbf{m}\right\},
\end{equation}
where $\theta_e$ is the encoder, $\theta_d$ is the decoder (discarded after pre-training), and $\mathbf{m}$ is the learned mask token together with the positional embeddings \citep{dosovitskiy2020image} (supportive parameters in the sense of Definition \ref{def_1}).

\textbf{The input perturbation.} Each image is cut into $L$ patches $\mathbf{x}_{n,1}, \dots, \mathbf{x}_{n,L}$, of which a fraction $\rho$ (in MAE, $\rho = 0.75$) is masked uniformly at random. Let $\mathcal{M}_n$ be the masked index set of image $n$, and $\mathcal{V}_n$ its complement. Then,
\begin{equation}\label{eq:mae_perturb}
    \mathcal{E}: \left\{\mathbf{X}_n\right\}_{n=1}^{N} \to
    \left\{\tilde{\mathbf{X}}_n\right\}_{n=1}^{N},
    \qquad
    \tilde{\mathbf{X}}_n \;=\;
    \Bigl(\, \left\{\mathbf{x}_{n,j}\right\}_{j \in \mathcal{V}_n},\;\;
    \mathcal{M}_n \,\Bigr).
\end{equation}
That is, each image is perturbed into its visible patches together with the record of which positions were removed. The encoder sees only $\left\{\mathbf{x}_{n,j}\right\}_{j \in \mathcal{V}_n}$, and the decoder receives the encoded visible tokens together with one copy of $\mathbf{m}$ per position in $\mathcal{M}_n$. The format of $\mathcal{E}$ in MAE is slightly more complicated than those in the above families, which still conforms to Definition \ref{def_1}.

\textbf{The training loss and the update.} The decoder predicts the pixels of the masked patches, and the loss is the mean squared error on those patches only, computed on (optionally per-patch normalized) pixel values:
\begin{equation}\label{eq:mae_loss}
    \mathcal{L}\!\left(\theta_e, \theta_d, \mathbf{m};\, \mathcal{E}(\mathcal{B})\right)
    \;=\; \frac{1}{\sum_n \left\vert \mathcal{M}_n \right\vert}
    \sum_{n=1}^{N} \sum_{j \in \mathcal{M}_n}
    \left\Vert \hat{\mathbf{x}}_{n,j} - \mathbf{x}_{n,j} \right\Vert_2^{2},
\end{equation}
where $\hat{\mathbf{x}}_{n,j}$ is the reconstruction of patch $j$. 

Again, all three components of $\Theta$ are updated in one gradient step,
\begin{equation}\label{eq:mae_omega}
    \Omega\!\left(\phi,\, \Theta,\, \mathcal{B},\, \mathcal{E}\right)
    \;=\; -\eta \left(\, \nabla_{\theta_e} \mathcal{L},\;\; \nabla_{\theta_d} \mathcal{L},\;\; \nabla_{\mathbf{m}} \mathcal{L} \,\right),
    \qquad
    \delta\Theta \;=\; \Omega\!\left(\phi,\, \Theta,\, \mathcal{B},\, \mathcal{E}\right) + \epsilon ,
\end{equation}
which obeys the format proposed in Definition \ref{def_1}. In summary, exactly one mask is drawn per image per step. There is no comparison among multiple masked candidates. Hence, \textbf{masked autoencoders are exploitative}. 

BEiT \citep{bao2021beit} replaces the pixel target by discrete visual tokens. The tokens come from a discrete variational autoencoder \citep{van2017neural}, which is trained beforehand and then frozen. So, its parameter has $\delta\theta=0$. The rest training process in BEiT is the same as that in MAE. Hence, BEiT is exploitative as well.

\subsection{Information-theoretic approaches}
\label{ssl_eval_info}

Finally, we evaluate information-theoretic approaches, taking IIC \citep{ji2019invariant} and VICReg \citep{bardes2021vicreg} as the two examples. In this family, the loss is not a sum over individual samples. Instead, the loss is a function of statistics aggregated over the whole batch, such as the joint distribution in IIC and the covariance matrix in VICReg. 

\textbf{The IIC model.} The parameter set is $\Theta = \{\theta\}$, where $\theta$ is the parameters of an encoder followed by a softmax clustering head, $f_\theta : \mathbb{R}^{H \times W \times C} \to [0,1]^{K}$.

Each image is augmented into two views. That is, $\mathcal{E} : \{\mathbf{X}_n\}_{n=1}^{N} \to \{(\tilde{\mathbf{X}}_n,\, \tilde{\mathbf{X}}'_n)\}_{n=1}^{N}$ (the way we describe $\mathcal{E}$ here differs from that in the Contrastive Learning sub-section \ref{ssl_eval_contrastive}, which is intended to easy the description in below. But both $\mathcal{E}$ are essentially the same, which augments one image into a pair of views). 

Using $\mathbf{p}_n = f_\theta(\tilde{\mathbf{X}}_n)$ and $\mathbf{p}'_n = f_\theta(\tilde{\mathbf{X}}'_n)$ to represent the two soft assignments, IIC forms the symmetrized joint distribution of the two views over the $K \times K$ grid of cluster pairs. In the meanwhile, the marginal distribution is simply a sum over matrix rows. That is,
\begin{equation}\label{eq:iic_eval_joint}
    P_{k,k'} \;=\; \frac{1}{2N}\sum_{n=1}^{N}
    \left(p_{n,k}\, p'_{n,k'} \;+\; p_{n,k'}\, p'_{n,k}\right),
    \qquad
    P_k \;=\; \sum\nolimits_{k'=1}^{K} P_{k,k'} .
\end{equation}
The training target is to maximize the mutual information between the two views' assignments:
\begin{equation}\label{eq:iic_eval_loss}
    I(P) \;=\; \sum_{k=1}^{K} \sum_{k'=1}^{K} P_{k,k'}
    \log \frac{P_{k,k'}}{P_k \, P_{k'}},
    \qquad
    \mathcal{L}\!\left(\theta;\, \mathcal{E}(\mathcal{B})\right) \;=\; -\,I(P).
\end{equation}
The matrix $P$ of Eq.~\ref{eq:iic_eval_joint} is a deterministic function of $\theta$ and $\mathcal{E}(\mathcal{B})$. The training target $I(P)$ is evaluated once on that single matrix. The parameter updates come from one gradient step,
\begin{equation}\label{eq:iic_eval_omega}
    \Omega\!\left(\phi,\, \Theta,\, \mathcal{B},\, \mathcal{E}\right) \;=\; -\eta\, \nabla_\theta \mathcal{L}\!\left(\theta;\, \mathcal{E}(\mathcal{B})\right),
    \qquad
    \delta\Theta \;=\; \Omega\!\left(\phi,\, \Theta,\, \mathcal{B},\, \mathcal{E}\right) + \epsilon .
\end{equation}
The full IIC model adds an auxiliary over-clustering head, which enlarges $\Theta$ by that head's parameters and adds a second loss of the same form as Eq.~\ref{eq:iic_eval_loss}. No comparison among candidates is introduced. Thus, the IIC model satisfies the conditions in Definition \ref{def_1}, \textbf{which is exploitative.}

\textbf{The VICReg model.} The parameter set is $\Theta = \{\theta\}$, where $\theta$ is the parameters of an encoder followed by an expander. Each image is again augmented into two views, $\mathcal{E} : \{\mathbf{X}_n\}_{n=1}^{N} \to \{(\tilde{\mathbf{X}}_n,\, \tilde{\mathbf{X}}'_n)\}_{n=1}^{N}$. The two views, in one batch of $N$ images, produce embedding matrices $Z, Z' \in \mathbb{R}^{N \times d}$, whose $n$-th rows are respectively $\mathbf{z}_n = f_\theta(\tilde{\mathbf{X}}_n)$ and $\mathbf{z}'_n = f_\theta(\tilde{\mathbf{X}}'_n)$. 

The loss has three terms: the invariance term pulls the two views of one image together; the variance term keeps every embedding dimension from collapsing, using a hinge at a target value $\gamma > 0$; and the covariance term de-correlates distinct embedding dimensions. That is,
\begin{equation}\label{eq:vicreg_terms}
\begin{aligned}
    s\!\left(Z, Z'\right) &= \frac{1}{N}\sum_{n=1}^{N} \left\Vert \mathbf{z}_n - \mathbf{z}'_n \right\Vert_2^{2},
    \qquad
    v(Z) = \frac{1}{d}\sum_{a=1}^{d} \max\!\left(0,\; \gamma - \sqrt{\mathrm{Var}\!\left(Z_{:,a}\right)}\right), \\
    c(Z) &= \frac{1}{d}\sum_{a \neq b} \left[\mathrm{Cov}(Z)\right]_{a,b}^{2}.
\end{aligned}
\end{equation}
With positive coefficients $\alpha_1, \alpha_2, \alpha_3$, the total loss is
\begin{equation}\label{eq:vicreg_loss}
    \mathcal{L}\!\left(\theta;\, \mathcal{E}(\mathcal{B})\right)
    \;=\; \alpha_1\, s\!\left(Z, Z'\right)
    \;+\; \alpha_2 \left(v(Z) + v(Z')\right)
    \;+\; \alpha_3 \left(c(Z) + c(Z')\right).
\end{equation}
Once again, the parameter updates come from one gradient step, with exactly the same form as Eq.~\ref{eq:iic_eval_omega}. \textbf{Hence, VICReg is exploitative.} 

Barlow Twins \citep{zbontar2021barlow} replaces the variance and covariance terms by driving the cross-correlation matrix between the two views toward the identity matrix. This changes the statistic being computed, but not the parameter updates process. Hence, Barlow Twins is exploitative as well.

One point deserves emphasis here, because it concerns our own approach. Our surprise score $\mathcal{S}(\theta)$ is also a batch-level statistic, and IIC's mutual information is its closest relative: Appendix \ref{connect_surprise_score_to_IIC} derives the exact connection between the two. However, $\mathcal{S}(\theta)$ cannot be fully optimized by exploitative approaches, due to its own particularity.

According to all the formulas and descriptions in Section \ref{sec:surprise}, we can see that each cluster (output dimension) in our model has its own `freedom' to either produce surprising view-matching result, or die out throughout the batch. That says, the final surprise score $\mathcal{S}(\theta)$ places no restriction on which clusters, or how many clusters, should finally produce surprising view-matching results. A cluster can either contribute to $\mathcal{S}(\theta)$, or die out quietly. With such a high freedom, there are astronomical possibilities for $\mathcal{S}(\theta)$ to select surprising clusters. Among these astronomical possibilities, the optimal one may not be achievable simply by exploiting observed states. Thus, explorative methods involving random perturbation and performance-based selection become necessary. But once again, this is an intuitive explanation, rather than a strict proof of Hypothesis \ref{hypothesis_1}.

Due to the overwhelming number of self-supervised learning approaches available today, it is impossible for us to evaluate every one of them in detail. Researchers can apply the evaluations above to the approaches they are interested in: First, write down the parameter set $\Theta$, the input perturbation $\mathcal{E}$ and the update $\delta\Theta$. Then, ask whether the fitness $\phi$ (or loss $\mathcal{L}$ in many cases) is ever evaluated at multiple $\Theta + \epsilon$ settings or multiple $\mathcal{E}(\mathcal{B})$ settings in order to decide $\delta\Theta$. 

In general, based on all six subsections above, we claim that all dominant SSL approaches for image clustering are exploitative.

\subsection{The two non-parametric baselines: DeepDPM and UNSEEN}
\label{ssl_eval_baselines}

Now, we evaluate the two baseline models used in Section \ref{experiments}: DeepDPM \citep{ronen2022deepdpm} and UNSEEN \citep{leiber2024dying}. To be honest, both approaches are non-parametric clustering algorithms applied to given features, whereas the features in both approaches are obtained separately. Note that neither approach augments input images. That is, $\mathcal{E} = \mathbf{I}$.

\textbf{The DeepDPM model.} DeepDPM performs inference in a Dirichlet Process Mixture (DPM) model \citep{neal2000markov} with a dynamic architecture. Its parameter set is
\begin{equation}
    \Theta \;=\; \Bigl\{\, \theta_{\mathrm{cl}},\;\;
    \bigl(\theta^{k}_{\mathrm{sub}}\bigr)_{k=1}^{K},\;\;
    \bigl(\pi_k,\, \boldsymbol{\mu}_k,\, \Sigma_k\bigr)_{k=1}^{K},\;\;
    \bigl(\tilde{\pi}_{k,j},\, \tilde{\boldsymbol{\mu}}_{k,j},\, \tilde{\Sigma}_{k,j}\bigr)_{k,\, j \in \{1,2\}} \,\Bigr\},
\end{equation}
where $\theta_{\mathrm{cl}}$ is a clustering network with $K$ output clusters, and $\theta^{k}_{\mathrm{sub}}$ is a sub-clustering network for cluster $k$ with two further output clusters (sub-clusters) that maintains a two-component split of cluster $k$. Then, $\bigl(\pi_k,\, \boldsymbol{\mu}_k,\, \Sigma_k\bigr)_{k=1}^{K}$ are respectively the weights, means and covariances of the $K$ main clusters; $\bigl(\tilde{\pi}_{k,j},\, \tilde{\boldsymbol{\mu}}_{k,j},\, \tilde{\Sigma}_{k,j}\bigr)_{k,\, j \in \{1,2\}}$ are those of the two sub-clusters in cluster $k$, indexed by $j \in \{1,2\}$.

For a fixed $K$, DeepDPM is exploitative. In each step, the clustering network takes one gradient step, with the loss to be the KL divergence \citep{shlens2014notes_KL_divergence_2} between the network's soft assignments and the assignments that an E-step in an EM algorithm \citep{dempster1977maximum} would produce from the current mixture parameters. The sub-clustering network takes one gradient step on an isotropic distance loss \citep{beckers1992optimization}. The mixture parameters are then updated by weighted maximum-a-posteriori estimates under a Normal-Inverse-Wishart prior \citep{so2024estimating}. All of these quantities are computed from the current $\Theta$ and the current batch $\mathcal{B}$ without performance-based selection or comparison. So, they are of the form $\Omega\!\left(\phi,\, \Theta,\, \mathcal{B},\, \mathcal{E}\right)$ required by Definition \ref{def_1}. 

Before continue, we note that this is a re-description of another model. One can refer to the original DeepDPM paper \citep{ronen2022deepdpm} for more details.

However, $K$ itself is changed by a mutation-selection mechanism. With every certain number of iterations, the split of each cluster $k$ into its two sub-clusters is proposed, and that proposal is accepted with probability $\min(1, H_{\mathrm{s}})$, where
\begin{equation}\label{eq:deepdpm_hastings}
    H_{\mathrm{s}} \;=\;
    \frac{\alpha\, \Gamma\!\left(N_{k,1}\right) f_x\!\left(X_{k,1}\right)
          \Gamma\!\left(N_{k,2}\right) f_x\!\left(X_{k,2}\right)}
         {\Gamma\!\left(N_k\right) f_x\!\left(X_k\right)}
\end{equation}
is the Metropolis--Hastings ratio \citep{hastings1970monte, chang2013parallel}. Here, $\alpha$ is the concentration parameter of the Dirichlet process \citep{teh2010dirichlet}, $\Gamma$ is the Gamma function, $X_k$ denotes the points in cluster $k$, $N_k = \left\vert X_k \right\vert$ denotes their amount, $X_{k,j}$ and $N_{k,j}$ are the same quantities for sub-cluster $j \in \{1,2\}$, and $f_x(\cdot)$ is the marginal likelihood under the Normal-Inverse-Wishart prior. Merge proposals are handled in the same way, with $H_{\mathrm{m}} = 1 / H_{\mathrm{s}}$, considered sequentially between each cluster and its three nearest neighbors. 

When a proposal is accepted, output units of the clustering network are added or removed, new sub-clustering networks are initialized, and $K$ changes accordingly. Once again, more details are provided in the original paper \citep{ronen2022deepdpm}.

As a result, DeepDPM randomly perturbs the number of main clusters $K$, and then decides whether to keep a perturbation based on the performance. Hence, \textbf{DeepDPM is explorative}.

It is worth noting where that exploration acts. DeepDPM directly performs on the output features of pre-trained vision models. In contrast, our converge-to-surprise scheme explores the entire parameter space of a network, which is trained from scratch on raw images.

\textbf{The UNSEEN framework.} UNSEEN is built upon existing deep clustering algorithms --- DCN \citep{yang2017towards}, DEC \citep{xie2016unsupervised} or DKM \citep{fard2020deep} --- and starts from an initial number of clusters $k_{\mathrm{init}}$ that acts as an upper bound. Using $C^{j}_{i}$ to denote the set of samples assigned to cluster $i$ in epoch $j$, the parameter set of UNSEEN is
\begin{equation}
    \Theta \;=\; \Bigl\{\, \theta_{\mathrm{enc}},\;\;
    \Theta_{\mathrm{base}},\;\; k^{j},\;\;
    \bigl(\left\vert C^{0}_{i} \right\vert\bigr)_{i} \,\Bigr\},
\end{equation}
where $\theta_{\mathrm{enc}}$ is the parameters of the encoder, $\Theta_{\mathrm{base}}$ collects the parameters of the clustering model built upon the encoder, $k^{j}$ is the number of active clusters in epoch $j$, and $\left\vert C^{0}_{i} \right\vert$ is the recorded creation size (initial size) of cluster $i$. The last two are supportive parameters: they are bookkeeping quantities that the update reads and writes.

The network is optimized by Adam \citep{AdamOptimization} on the clustering algorithm's own loss, together with UNSEEN's nearest-neighbor term:
\begin{equation}\label{eq:unseen_loss}
    \mathcal{L}_{\mathrm{UNSEEN}}\!\left(\mathcal{B},\, l\right)
    \;=\; \frac{1}{l \left\vert \mathcal{B} \right\vert}
    \sum_{\mathbf{X} \in \mathcal{B}} \;
    \sum_{\mathbf{Y} \in \mathrm{nn}\left(\mathcal{B},\, \mathbf{X},\, l\right)}
    \left\Vert \mathrm{enc}(\mathbf{X}) - \mathrm{enc}(\mathbf{Y}) \right\Vert_2^{2}.
\end{equation}
Such a loss pulls each sample toward its $l$ nearest neighbors within the batch, so that samples can still change clusters late in training. Note that the encoder is trained from scratch with freshly initialized $\theta_{\mathrm{enc}}$. So, unlike DeepDPM, UNSEEN does not apply pre-trained models.

Then, after each epoch, cluster $i$ is declared dead whenever its current size has shrunk below a fraction $t$ of its creation size. That is, if
\begin{equation}\label{eq:unseen_dying}
    \left\vert C^{j}_{i} \right\vert/ \left\vert C^{0}_{i} \right\vert \;<\; t ,
    \qquad t = 0.5 \;\;\text{by default},
\end{equation}
then cluster $i$ will be removed completely. Its samples are reassigned to the closest remaining cluster. The creation sizes of the receiving clusters are increased by the number of samples they absorb, and $k^{j}$ is set to the number of survivors. Eq.~\ref{eq:unseen_dying} is the mechanism by which the number of clusters is inferred. There is no split operation. Once again, more details can be discovered from the original paper \citep{leiber2024dying}.

To write the parameter updates explicitly, we first fix the notation of the assignment. Let $\mathcal{X}$ be the dataset, and let $A^{j}$ be the index set of active clusters in epoch $j$ (so $k^{j} = \left\vert A^{j} \right\vert$). Let $a^{j} : \mathcal{X} \to A^{j}$ be the hard assignment produced by the clustering algorithm, so that $C^{j}_{i} = \{\mathbf{X} \in \mathcal{X} : a^{j}(\mathbf{X}) = i\}$. Then, Eq.~\ref{eq:unseen_dying} defines the set of dead clusters and the set of survivors,
\begin{equation}\label{eq:unseen_dead_set}
    D^{j} \;=\; \Bigl\{\, i \in A^{j} \;:\;
    \left\vert C^{j}_{i} \right\vert / \left\vert C^{0}_{i} \right\vert < t \,\Bigr\},
    \qquad
    A^{j+1} \;=\; A^{j} \setminus D^{j} .
\end{equation}
Every sample of a dead cluster is then reassigned by the same rule restricted to the survivors, $\hat{a}^{j} : \mathcal{X} \to A^{j+1}$. We write
\begin{equation}\label{eq:unseen_absorbed}
    n^{j}_{i} \;=\; \Bigl\vert \Bigl\{\, \mathbf{X} \in
    \bigcup\nolimits_{i' \in D^{j}} C^{j}_{i'} \;:\;
    \hat{a}^{j}(\mathbf{X}) = i \,\Bigr\} \Bigr\vert ,
    \qquad i \in A^{j+1},
\end{equation}
to denote the number of samples that the surviving cluster $i$ absorbs. Note that $\sum_{i \in A^{j+1}} n^{j}_{i} = \sum_{i' \in D^{j}} \left\vert C^{j}_{i'} \right\vert$: no sample is lost.

With $\mathcal{L} = \mathcal{L}_{\mathrm{base}} + \mathcal{L}_{\mathrm{UNSEEN}}$ (again, UNSEEN is built upon existing deep clustering algorithms. So, $\mathcal{L}_{\mathrm{base}}$ is the loss of that base model), the updates of the four components of $\Theta$ come in two stages: 

Within epoch $j$, each batch produces one gradient step and touches nothing else (of course there are multiple steps in one batch):
\begin{equation}\label{eq:unseen_update_grad}
    \delta\theta_{\mathrm{enc}} = -\eta\, \nabla_{\theta_{\mathrm{enc}}} \mathcal{L}\!\left(\Theta;\, \mathcal{B}\right),
    \qquad
    \delta\Theta_{\mathrm{base}} = -\eta\, \nabla_{\Theta_{\mathrm{base}}} \mathcal{L}\!\left(\Theta;\, \mathcal{B}\right),
    \qquad
    \delta k^{j} = 0,
    \qquad
    \delta \left\vert C^{0}_{i} \right\vert = 0 .
\end{equation}
At the end of epoch $j$, the pruning stage applies Eq.~\ref{eq:unseen_dead_set}--\ref{eq:unseen_absorbed}:
\begin{equation}\label{eq:unseen_update_prune}
    \delta\theta_{\mathrm{enc}} = 0,
    \qquad
    \delta\Theta_{\mathrm{base}} = \Pi_{A^{j+1}} \Theta_{\mathrm{base}} - \Theta_{\mathrm{base}},
    \qquad
    \delta k^{j} = -\left\vert D^{j} \right\vert,
    \qquad
    \delta \left\vert C^{0}_{i} \right\vert = n^{j}_{i}
    \;\;\left(i \in A^{j+1}\right),
\end{equation}
where $\Pi_{A^{j+1}}$ keeps only the entries indexed by the surviving clusters, so that the cluster parameters of every $i \in D^{j}$ --- together with their recorded creation sizes --- are deleted from $\Theta$. Two features of Eq.~\ref{eq:unseen_update_prune} are worth naming. The count $k^{j}$ decreases by exactly the number of dead clusters and never increases, since there is no split operation. And a creation size never tracks the current size of its cluster: by Eq.~\ref{eq:unseen_absorbed}, $\left\vert C^{0}_{i} \right\vert$ changes only when cluster $i$ absorbs samples from a dead cluster, and is otherwise frozen at the value it was given when the cluster was created.

Every quantity in Eq.~\ref{eq:unseen_update_grad}--\ref{eq:unseen_update_prune} is determined by the current parameter set $\Theta$ and the current data. The dead set $D^{j}$ is read off the counts $\left\vert C^{j}_{i} \right\vert$ and the recorded $\left\vert C^{0}_{i} \right\vert$. The absorbed counts $n^{j}_{i}$ come from the clustering algorithm's own assignment rule. In particular, Eq.~\ref{eq:unseen_dying} is a threshold evaluated at the single current $\Theta$. 

Collecting Eq.~\ref{eq:unseen_update_grad}--\ref{eq:unseen_update_prune} into $\Omega$, we have \begin{equation}\label{eq:UNSEEN_final}
    \Omega\!\left(\phi,\, \Theta,\, \mathcal{B},\, \mathcal{E}\right)=(\delta\theta_{\mathrm{enc}},\;\;
    \delta\Theta_{\mathrm{base}},\;\; \delta k^{j},\;\;
    \bigl(\delta\left\vert C^{0}_{i} \right\vert\bigr)_{i}), \ \ \ \ \ \ \ \delta\Theta = \Omega\!\left(\phi,\, \Theta,\, \mathcal{B},\, \mathcal{E}\right) + \epsilon.
\end{equation}

One minor issue is that, the fitness score $\phi$ of $k^j$ and $\bigl(\left\vert C^{0}_{i} \right\vert\bigr)_{i}$ is not explicitly defined. But since we can explicitly describe Eq.~\ref{eq:UNSEEN_final}, we do not need to be bothered by this minor issue. In summary, UNSEEN never instantiates or scores a second parameter set, and never keeps one candidate over another according to performances. Hence, \textbf{UNSEEN is exploitative}.

The contrast between these two baselines is informative. Both infer the number of clusters, both operate on a learned embedding, and both are trained without labels. Yet the explorative one, DeepDPM, outperforms all three exploitative UNSEEN variants on every one of the three benchmarks in Table \ref{tab:main-results}. Our converge-to-surprise scheme is explorative as well, which explores the full parameter set of a network and outperforms DeepDPM on most metrics in Table \ref{tab:main-results}. This may be regarded as an indirect supportive evidence of Hypothesis \ref{hypothesis_1}: On the same benchmarks with the same non-parametric clustering tasks, explorative approaches in general performs better than exploitative ones, indicating that explorative approaches are better at `exploring' the parameter space to find an optimum. Hence, explorative approaches should also be better at optimizing the surprise score, which requires exploring the deep network's parameter space. But once again, this is another intuitive analysis.

\textbf{Remark (scope of Definition \ref{def_1}).} The following sub-sections no more deal with discriminative learning on images. Thus, some assumptions in Definition \ref{def_1} need to be expanded or modified. 

First, we allow $\mathbf{P}$ to be a distribution over an arbitrary sample space. 

Second, a single fitness score $\phi$ is not required. When an approach partitions $\Theta$ into blocks and updates separately, each block carries its own fitness score (or scores). 

Third, $\epsilon$ and $\mathcal{E}$ stand for whatever random factors the approach draws, and need not be the only two. The criterion itself is unchanged: the approach is exploitative when no update is decided by evaluating fitness scores at multiple settings of those random factors, and explorative otherwise.

\subsection{Expanding our scope: generative adversarial networks}
\label{ssl_eval_gan}

Every approach evaluated above is discriminative \citep{jebara2012machine}: one parameter set is trained toward one fitness (explicit or implicit). Generative adversarial networks (GAN) \citep{goodfellow2014generative} differ in two aspects. They are generative models, and they are trained by a min-max game between two parameter blocks with opposing objectives. %So Definition \ref{def_1} does not apply verbatim. It applies \emph{block-wise}: we partition $\Theta$ into the blocks that the approach updates separately, give each block its own fitness, and then ask the question of Definition \ref{def_1} once per block --- is that block's update decided by comparing its fitness across multiple candidate settings? If no block is updated in that manner, the approach as a whole is exploitative.

We take the original GAN of \citep{goodfellow2014generative} as the example. Its parameter set is
\begin{equation}
    \Theta \;=\; \left\{\theta_g,\; \theta_d\right\},
\end{equation}
where $\theta_g$ is the parameters of a generator $G_{\theta_g} : \mathbb{R}^{d_z} \to \mathbb{R}^{H \times W \times C}$, which maps a latent code to an image, and $\theta_d$ is the parameters of a discriminator $D_{\theta_d} : \mathbb{R}^{H \times W \times C} \to [0,1]$, which estimates the probability that its input is a real image rather than a generated one. We keep the standard symbols $G$ and $D$ here. The discriminator $D_{\theta_d}(\cdot)$ takes a single argument, and is not to be confused with the binary Kullback--Leibler divergence $D(\cdot \Vert \cdot)$ of Eq.~\ref{eq:surprise}.

\textbf{The input perturbation.} The original GAN applies no augmentation to the real images. Its random factor is the latent code instead: $N$ codes are drawn independently from a fixed prior $p_z$, such as a Gaussian, and passed to the generator. This is the perturbation of Definition \ref{def_1}, which attaches one code to each sample of the batch:
\begin{equation}\label{eq:gan_perturb}
    \mathcal{E}: \left\{\mathbf{X}_n\right\}_{n=1}^{N} \to
    \left\{\left(\mathbf{X}_n,\, \mathbf{z}_n\right)\right\}_{n=1}^{N},
    \qquad \mathbf{z}_n \sim p_z .
\end{equation}

\textbf{The training target.} The two blocks (generator and discriminator) share a single value function, which the discriminator maximizes (ascends) and the generator minimizes (descends):
\begin{equation}\label{eq:gan_value}
    \min_{\theta_g} \max_{\theta_d} \; V\!\left(\theta_d,\, \theta_g\right)
    \;=\; \mathbb{E}_{\mathbf{X} \sim \mathbf{P}}
    \left[\log D_{\theta_d}(\mathbf{X})\right]
    \;+\; \mathbb{E}_{\mathbf{z} \sim p_z}
    \left[\log\!\left(1 - D_{\theta_d}\!\left(G_{\theta_g}(\mathbf{z})\right)\right)\right].
\end{equation}
Read block-wise, the fitness of the discriminator is $\phi_d = V$ and the fitness of the generator is $\phi_g = -V$.

\textbf{The updates.} Training alternates between a fixed number of ascent steps on $\theta_d$ and one descent step on $\theta_g$ over the current batch:
\begin{equation}\label{eq:gan_updates}
\begin{aligned}
    \delta\theta_d &= \eta\, \nabla_{\theta_d} \frac{1}{N}\sum_{n=1}^{N}
    \left[\log D_{\theta_d}(\mathbf{X}_n)
    + \log\!\left(1 - D_{\theta_d}\!\left(G_{\theta_g}(\mathbf{z}_n)\right)\right)\right], \\
    \delta\theta_g &= -\eta\, \nabla_{\theta_g} \frac{1}{N}\sum_{n=1}^{N}
    \log\!\left(1 - D_{\theta_d}\!\left(G_{\theta_g}(\mathbf{z}_n)\right)\right).
\end{aligned}
\end{equation}
In practice, the generator maximizes $\log D_{\theta_d}(G_{\theta_g}(\mathbf{z}_n))$ instead, which avoids vanishing gradients early in training. This changes the gradient, but not the structure of the update. Collecting the two blocks, we have that
\begin{equation}\label{eq:gan_omega}
    \Omega\!\left(\phi,\, \Theta,\, \mathcal{B},\, \mathcal{E}\right)
    \;=\; \left(\delta\theta_g,\; \delta\theta_d\right),
    \qquad
    \delta\Theta \;=\; \Omega\!\left(\phi,\, \Theta,\, \mathcal{B},\, \mathcal{E}\right) + \epsilon .
\end{equation}
Both entries are computed once, at the current $\left(\theta_g, \theta_d\right)$ and on the single perturbed batch $\mathcal{E}(\mathcal{B})$ of Eq.~\ref{eq:gan_perturb}. Hence, \textbf{GAN is exploitative}.

The min-max game may look like an explorative process, so we clarify it here. Both blocks in Eq.~\ref{eq:gan_updates} are updated by gradient operators, applied unconditionally and with opposite signs. Neither block ever proposes a candidate parameter setting, evaluates $V$ at it, and then keeps or discards it according to the evaluation. The discriminator does score the generated samples, but that is through a direct training signal rather than an acceptance test. After the discriminator ascends the fitness $V$, the generator descends it, whether or not the step improves the ability of the generator. In contrast, the centered rank $r_i$ in Eq.~\ref{eq:es-update} of our converge-to-surprise scheme multiplies the perturbation $\epsilon_i$ itself, so there are performance-based selection (ranking) on perturbations in each optimization step. In summary, the min-max game in GAN does not involve a mutation-selection mechanism.

\subsection{Expanding our scope: diffusion models}
\label{ssl_eval_diffusion}

Diffusion models are generative as well, but they are trained without an adversary. So, Definition \ref{def_1} applies to them directly. We take the denoising diffusion probabilistic model (DDPM) of \citep{ho2020denoising} as the example. Its parameter set is simply
\begin{equation}
    \Theta \;=\; \left\{\theta\right\}.
\end{equation}
Here, $\theta$ is the parameters of a network $\boldsymbol{\xi}_\theta\!\left(\cdot,\, t\right) : \mathbb{R}^{H \times W \times C} \to \mathbb{R}^{H \times W \times C}$, which receives a noised image together with its noise level $t$, and predicts the noise that was added to it. Here, we use $\boldsymbol{\xi}$ to denote the Gaussian noise, rather than using $\boldsymbol{\epsilon}$ of \citep{ho2020denoising}, since $\epsilon$ already denotes the parameter-level perturbation in Definition \ref{def_1}. One will not confuse $\boldsymbol{\xi}$ with the network symbol $\boldsymbol{\xi}_\theta$.

\textbf{The input perturbation.} The random factor of a diffusion model is the forward (noising/diffusion) process, which is applied to the batch and is itself the pretext task. For each image, a time step $t_n$ is drawn uniformly from $\{1, \dots, T\}$, and a noise tensor $\boldsymbol{\xi}_n$ is drawn from a standard Gaussian. Then, the image is corrupted accordingly:
\begin{equation}\label{eq:ddpm_perturb}
    \mathcal{E}: \left\{\mathbf{X}_n\right\}_{n=1}^{N} \to
    \left\{\left(\tilde{\mathbf{X}}_n,\, t_n,\, \boldsymbol{\xi}_n\right)\right\}_{n=1}^{N},
    \qquad
    \tilde{\mathbf{X}}_n = \sqrt{\bar{\alpha}_{t_n}}\, \mathbf{X}_n
    + \sqrt{1 - \bar{\alpha}_{t_n}}\; \boldsymbol{\xi}_n .
\end{equation}
Here, $t_n \sim \mathrm{Uniform}\{1, \dots, T\}$, $\boldsymbol{\xi}_n \sim \mathcal{N}(0, I)$, and $\bar{\alpha}_t \in (0,1)$ is a fixed decreasing variance schedule, which is not learned. Eq.~\ref{eq:ddpm_perturb} plays the same role as the random masking in MAE (Eq.~\ref{eq:mae_perturb}): a random rule destroys part of the information in each sample, and the model is trained to undo it.

\textbf{The training target and the update.} The network is trained to predict the noise that Eq.~\ref{eq:ddpm_perturb} added, under the simplified objective of \citep{ho2020denoising}:
\begin{equation}\label{eq:ddpm_loss}
    \mathcal{L}\!\left(\theta;\, \mathcal{E}(\mathcal{B})\right)
    \;=\; \frac{1}{N}\sum_{n=1}^{N}
    \left\Vert \boldsymbol{\xi}_n - \boldsymbol{\xi}_\theta\!\left(\tilde{\mathbf{X}}_n,\, t_n\right) \right\Vert_2^{2} .
\end{equation}
Similar as in the discriminative models, the fitness is $\phi=\mathcal{L}\!\left(\theta;\, \mathcal{E}(\mathcal{B})\right)$. The parameter updates come from one gradient step,
\begin{equation}\label{eq:ddpm_omega}
    \Omega\!\left(\phi,\, \Theta,\, \mathcal{B},\, \mathcal{E}\right)
    \;=\; -\eta\, \nabla_{\theta} \mathcal{L}\!\left(\theta;\, \mathcal{E}(\mathcal{B})\right),
    \qquad
    \delta\Theta \;=\; \Omega\!\left(\phi,\, \Theta,\, \mathcal{B},\, \mathcal{E}\right) + \epsilon .
\end{equation}
Exactly one perturbation pair $\left(t_n, \boldsymbol{\xi}_n\right)$ is drawn per image per step, and it is directly applied to the sample image without performance-based selection. Hence, \textbf{diffusion models (DDPMs) are exploitative}.

During inference, the iterative reverse process may look like an explorative process, so we clarify it here. Starting from pure noise, $\boldsymbol{\xi}_\theta$ is applied for $T$ steps to produce an image. There is no parameter update at all, which is therefore outside the scope of Definition \ref{def_1}. Later diffusion models, such as the transformer-based DiT \citep{peebles2023scalable}, change the architecture and the space in which the diffusion is run, but leave the fundamental denoising mechanism unchanged.

DDIM \citep{song2020denoising} differs from DDPM in the reverse process rather than in the training. It replaces the Markovian reverse chain by a family of non-Markovian ones that share the same forward marginals as Eq.~\ref{eq:ddpm_perturb}, so the training target remains Eq.~\ref{eq:ddpm_loss} and a DDPM-trained network can be reused unchanged. What DDIM gains is at inference time: the sampling becomes deterministic given the initial noise, and it may jump along a subsequence of the $T$ steps, which is considerably faster. Once again, there is no performance-based selection. The noise is applied directly to the samples, and all parameter updates come from one gradient step. Hence, DDIMs are also exploitative.

\subsection{Expanding our scope: reinforcement learning}
\label{ssl_eval_rl}

Reinforcement learning (RL) \citep{sutton1998reinforcement} needs two further relaxations of the setting in Definition \ref{def_1}. First, the samples are not drawn from a distribution that is fixed and independent of $\Theta$: the agent generates its own data, so the sample distribution follows the current parameters. Second, the random factor is neither a perturbation $\mathcal{E}$ of a given batch nor a perturbation $\epsilon$ of the parameters, but the agent's own random choice of action. The criterion of Definition \ref{def_1} is unchanged: we ask whether the fitness is evaluated at multiple settings of that random factor in order to decide $\delta\Theta$.

We take tabular Q-learning \citep{watkins1992q} as the example, which is among the simplest RL approaches. Its parameter set is
\begin{equation}
    \Theta \;=\; \left\{\theta_{Q}\right\},
    \qquad
    Q : \mathcal{S} \times \mathcal{A} \to \mathbb{R},
\end{equation}
where $\mathcal{S}$ is the set of states, $\mathcal{A}$ is the set of actions, $Q(s, a)$ is the current estimate of the value of taking action $a$ in state $s$, and $\theta_{Q}$ is the parameters of the deep Q-learning network.

\textbf{The random factor.} In state $s_{t}$, the agent selects an action $a_{t}$ using an $\epsilon$-greedy behavior policy: it chooses a uniformly random action with a small, fixed probability $\epsilon$, and selects the greedy action (the one with the highest current value) otherwise. The environment then returns a reward $R_t$ and a next state $s_{t+1}$. The transition $\left(s_t,\, a_t,\, R_t,\, s_{t+1}\right)$ is the sample, and it exists only because that random choice was made.

\textbf{The update.} With learning rate $\eta$ and discount factor $\gamma \in [0, 1)$, only the entry of the action that was actually taken is updated:
\begin{equation}\label{eq:qlearning_update}
    \delta Q\!\left(s_t,\, a_t\right)
    \;=\; \eta \Bigl[\, R_t \;+\; \gamma \max_{a' \in \mathcal{A}} Q\!\left(s_{t+1},\, a'\right)
    \;-\; Q\!\left(s_t,\, a_t\right) \Bigr] ,
\end{equation}
while $\delta Q(s, a) = 0$ for every other state-action pair.

Eq.~\ref{eq:qlearning_update} applies performance-based selection to a random factor. The bracketed term is the outcome of the randomly chosen action $a_t$, measured against the value that was expected of it, and the value of exactly that action is then updated. If the random choice turned out better than expected, its value is raised, and the behavior policy will take it more often; if it turned out worse, its value is lowered and the choice is suppressed. So, the random factor is retained or discarded according to the performance it produced, and the samples of the following steps are generated accordingly. This means that we cannot find a map $\Omega$, with only directly applied perturbations, to represent parameter updates in reinforcement learning. Hence, \textbf{reinforcement learning is explorative}.

Once again, there is a wide range of self-supervised learning approaches: generative, discriminative and behavioral. We cannot evaluate each and every of them. Researchers can apply similar evaluations to the approaches they are interested in.

%The parallel with our converge-to-surprise scheme is close. In Eq.~\ref{eq:es-update}, the centered rank $r_i$ multiplies the perturbation $\epsilon_i$ itself, so each perturbation is kept in proportion to how it ranked among the population. In Eq.~\ref{eq:qlearning_update}, the bracketed term plays the same role for the action that was sampled. Policy gradient approaches \citep{williams1992bizarreness} make the correspondence even more explicit: there, the update is the gradient of the log-probability of the sampled action, weighted by the return that this action achieved. It is therefore no coincidence that the evolution strategy we adopt was proposed as an alternative to reinforcement learning \citep{salimans2017evolution}.

\section{Connection between our surprise score $\mathcal{S}(\theta)$ and the IIC model training target}
\label{connect_surprise_score_to_IIC}

Recall from Section~\ref{sec:surprise} that, given $N$ images, the two complementary views of image $n$ produce logit vectors $\mathbf{y}_n^{(i)} = f_\theta(\tilde{\mathbf{X}}_n^{(i)})$ and $\mathbf{y}_n^{(j)} = f_\theta(\tilde{\mathbf{X}}_n^{(j)})$, from which the hard assignments $\hat y_n^{(i)}, \hat y_n^{(j)}$ are taken by argmax. But now, if we replace each argmax by a softmax at temperature $\tau'>0$, we will obtain soft assignments on the probability simplex,
\begin{equation}
    \boldsymbol{\pi}_n = \mathrm{softmax}\!\left(\mathbf{y}_n^{(i)} / \tau'\right),
    \qquad
    \boldsymbol{\rho}_n = \mathrm{softmax}\!\left(\mathbf{y}_n^{(j)} / \tau'\right).
\end{equation}
Before continue, note that one should not confuse the temperature $\tau'$ with the surprising cluster threshold $\tau$ used in formula \ref{eq:contributing} of our main paper.

Then, every count in Section~\ref{sec:surprise} becomes a soft expectation:
\begin{equation}
    \hat q_k = \frac{1}{N}\sum_{n=1}^{N} \pi_{n,k}\,\rho_{n,k},
    \qquad
    p_k = \frac{1}{2N}\sum_{n=1}^{N}\left(\pi_{n,k} + \rho_{n,k}\right),
    \qquad
    q_k = p_k^{\,2}.
\end{equation}
Also, the over-matching indicator $\mathbf{1}[\hat q_k > q_k]$ of Eq.~\ref{eq:surprise} can be replaced by a smooth gate, namely the logistic sigmoid \citep{kyurkchiev2015_sigmoid}:
\begin{equation}\label{eq:smooth_gate}
    \sigma(u) \;=\; \frac{1}{1 + e^{-u}}, \qquad u \in \mathbb{R}.
\end{equation}
By a routine derivation, we can prove that
\begin{equation}\label{eq:gate_limit}
    \lim_{\varepsilon \to 0^{+}} \sigma\!\left(\frac{\hat q_k - q_k}{\varepsilon}\right) \;=\; \mathbf{1}\!\left[\hat q_k > q_k\right].
\end{equation}
So, the hard selection of over-matching clusters is recovered as $\varepsilon \to 0^{+}$. Then, we define the soft version of the surprise score as
\begin{equation}\label{eq:soft_surprise}
    \mathcal{S}_{\mathrm{soft}}(\theta;\tau',\varepsilon)
    = \sum\nolimits_{k} \sigma\!\left(\frac{\hat q_k - q_k}{\varepsilon}\right) D\!\left(\hat q_k \,\Vert\, q_k\right).
\end{equation}
Again, by a routine derivation, we can prove that
\begin{equation}
    \lim_{\tau',\varepsilon \to 0^+} \mathcal{S}_{\mathrm{soft}}(\theta;\tau',\varepsilon) = \mathcal{S}(\theta).
\end{equation}
$\mathcal{S}_{\mathrm{soft}}$ is differentiable in $\theta$, so it can be maximized directly by Adam \citep{AdamOptimization}. This is worth stating plainly: we do not claim that the surprise score cannot be made differentiable. It can, and Eq.~\ref{eq:soft_surprise} is one way to do it.

Now, we compare the above format with the training target of the IIC model. Note that the IIC model does not apply chessboard masking as in our approach. The IIC model directly augment the original image into two views. But the output of the model on the two views of image $n$ can be represented by $\boldsymbol{\pi}_n$ and $\boldsymbol{\rho}_n$ all the same. 

Then, given the same soft assignments $\boldsymbol{\pi}_n$ and $\boldsymbol{\rho}_n$, IIC forms the soft joint distribution of the two views' cluster assignments over the $K \times K$ grid of assignment pairs:
\begin{equation}\label{eq:iic_joint_raw}
    \tilde P_{k,k'} \;=\; \frac{1}{N}\sum_{n=1}^{N} \pi_{n,k}\,\rho_{n,k'} .
\end{equation}
Entry $(k,k')$ is the soft frequency with which the $i$-side view is assigned to cluster $k$ while the $j$-side view is assigned to cluster $k'$. IIC then symmetrizes this matrix,
\begin{equation}\label{eq:iic_joint}
    P \;=\; \tfrac{1}{2}\left(\tilde P + \tilde P^{\!\top}\right),
    \qquad\text{that is,}\qquad
    P_{k,k'} \;=\; \frac{1}{2N}\sum_{n=1}^{N}\left(\pi_{n,k}\,\rho_{n,k'} + \pi_{n,k'}\,\rho_{n,k}\right).
\end{equation}
That is, the IIC model training target encodes the fact that the two augmented views play interchangeable roles. Neither view is ``the original'' and neither is ``the transformed one''. This is exactly as in our own chessboard-masking construction, where the $i$-side and the $j$-side are two halves of one chessboard with no side being original.

We now read off the two quantities of Section~\ref{sec:surprise} from this matrix.

\textbf{The diagonal.} Setting $k' = k$ in Eq.~\ref{eq:iic_joint}, we have:
\begin{equation}\label{eq:diag_derivation}
    P_{k,k}
    \;=\; \frac{1}{2N}\sum_{n=1}^{N}\left(\pi_{n,k}\,\rho_{n,k} + \pi_{n,k}\,\rho_{n,k}\right)
    \;=\; \frac{1}{N}\sum_{n=1}^{N}\pi_{n,k}\,\rho_{n,k}
    \;=\; \hat q_k .
\end{equation}
So, the $k$-th diagonal entry of IIC's joint distribution equals to our soft view-matching rate: the soft probability that both views of the same image are clustered to $k$.

\textbf{The marginal.} Summing the $k$-th row of Eq.~\ref{eq:iic_joint} over $k'$ recovers our marginal cluster probability $p_k$. That is,
\begin{align}
    P_k &\;=\; \sum_{k'} P_{k,k'}\label{eq:marg_0}\\[2pt]
    &\;=\; \sum_{k'} \frac{1}{2N}\sum_{n=1}^{N}\left(\pi_{n,k}\,\rho_{n,k'} + \pi_{n,k'}\,\rho_{n,k}\right)
    \label{eq:marg_1}\\[2pt]
    &\;=\; \frac{1}{2N}\sum_{n=1}^{N}\sum_{k'}\left(\pi_{n,k}\,\rho_{n,k'} + \pi_{n,k'}\,\rho_{n,k}\right)
    \label{eq:marg_2}\\[2pt]
    &\;=\; \frac{1}{2N}\sum_{n=1}^{N}\left(\pi_{n,k}\sum_{k'}\rho_{n,k'} \;+\; \rho_{n,k}\sum_{k'}\pi_{n,k'}\right)
    \label{eq:marg_3}\\[2pt]
    &\;=\; \frac{1}{2N}\sum_{n=1}^{N}\left(\pi_{n,k}\cdot 1 \;+\; \rho_{n,k}\cdot 1\right)
    \label{eq:marg_4}\\[2pt]
    &\;=\; \frac{1}{2N}\sum_{n=1}^{N}\left(\pi_{n,k}+\rho_{n,k}\right) \;=\; p_k .
    \label{eq:marg_5}
\end{align}
The step from Eq.~\ref{eq:marg_3} to Eq.~\ref{eq:marg_4} is true because
\begin{equation}\label{eq:simplex}
    \sum\nolimits_{k'} \pi_{n,k'} \;=\; \sum\nolimits_{k'} \rho_{n,k'} \;=\; 1 .
\end{equation}

It is worth noting what would happen without the symmetrization of Eq.~\ref{eq:iic_joint}. Running the same computation on the raw joint of Eq.~\ref{eq:iic_joint_raw} gives $\sum_{k'} \tilde P_{k,k'} = \tfrac{1}{N}\sum_{n} \pi_{n,k}$ and $\sum_{k} \tilde P_{k,k'} = \tfrac{1}{N}\sum_{n} \rho_{n,k'}$. That says, the default symmetrization in both the IIC model and our approach ensures that the joint and marginal probabilities above are equal.

Collecting Eq.~\ref{eq:diag_derivation} and Eq.~\ref{eq:marg_5}, we obtain the correspondence
\begin{equation}\label{eq:correspondence}
    P_{k,k} = \frac{1}{N}\sum_{n=1}^{N}\pi_{n,k}\,\rho_{n,k} = \hat q_k,
    \qquad
    P_k = \sum\nolimits_{k'} P_{k,k'} = \frac{1}{2N}\sum_{n=1}^{N}\left(\pi_{n,k}+\rho_{n,k}\right) = p_k .
\end{equation}
That is, our observed view-matching rate $\hat q_k$ equals to the $k$-th diagonal entry of IIC's joint distribution, and our marginal distribution also equals to theirs.

Finally, the training target of the IIC model is the mutual information between the two views' assignments:
\begin{equation}\label{eq:iic_mi}
    I(\boldsymbol{\pi};\boldsymbol{\rho}) \;=\; \sum\nolimits_{k,k'} P_{k,k'} \log \frac{P_{k,k'}}{P_k\, P_{k'}} .
\end{equation}

Now, we compare the two training targets directly. Removing the smooth gate from Eq.~\ref{eq:soft_surprise} leaves the un-gated soft surprise score, which no longer depends on $\varepsilon$:
\begin{equation}\label{eq:ungated}
    \tilde{\mathcal{S}}_{\mathrm{soft}}(\theta;\tau')
    \;=\; \sum\nolimits_{k} D\!\left(\hat q_k \,\Vert\, q_k\right).
\end{equation}
Substituting the correspondence of Eq.~\ref{eq:correspondence} into Eq.~\ref{eq:ungated}, and subtracting Eq.~\ref{eq:iic_mi}, we have:
\begin{equation}\label{eq:soft_minus_mi}
\begin{aligned}
    &\tilde{\mathcal{S}}_{\mathrm{soft}}(\theta;\tau') - I(\boldsymbol{\pi};\boldsymbol{\rho})
    = \sum_{k} D\!\left(\hat q_k \,\Vert\, q_k\right)
       \;-\; \sum_{k,k'} P_{k,k'}\log\frac{P_{k,k'}}{P_k P_{k'}} \\[4pt]
    &= \sum_{k}\left[\hat q_k\log\frac{\hat q_k}{q_k}
       + \left(1-\hat q_k\right)\log\frac{1-\hat q_k}{1-q_k}\right]
       \;-\; \sum_{k,k'} P_{k,k'}\log\frac{P_{k,k'}}{P_k P_{k'}} \\[4pt]
    &= \sum_{k}\left[P_{k,k}\log\frac{P_{k,k}}{P_k P_k}
       + \left(1-P_{k,k}\right)\log\frac{1-P_{k,k}}{1-P_k P_k}\right]
       \;-\; \sum_{k,k'} P_{k,k'}\log\frac{P_{k,k'}}{P_k P_{k'}} \\[4pt]
    &= \sum_{k}P_{k,k}\log\frac{P_{k,k}}{P_k P_k} + \sum_{k} \left(1-P_{k,k}\right)\log\frac{1-P_{k,k}}{1-P_k P_k}
       - \sum_{k}P_{k,k}\log\frac{P_{k,k}}{P_k P_k}
       - \sum_{k \neq k'} P_{k,k'}\log\frac{P_{k,k'}}{P_k P_{k'}} \\[4pt]
    &= \sum_{k} \left(1-P_{k,k}\right)\log\frac{1-P_{k,k}}{1-P_k P_k} \;-\; \sum_{k \neq k'} P_{k,k'}\log\frac{P_{k,k'}}{P_k P_{k'}} .
\end{aligned}
\end{equation}

For simplicity, we write $R_k=\left(1-P_{k,k}\right)\log\frac{1-P_{k,k}}{1-P_k P_k}$. Here is our intuitive analysis on Eq.~\ref{eq:soft_minus_mi}:

\begin{enumerate}[leftmargin=*]
    \item The diagonal sum cancels between the two targets. That is, on the diagonal, our surprise score (without considering over-matching or under-matching) and the IIC model training target agree exactly. This is the precise sense in which our differentiable surprise score is the diagonal of the IIC model training target. Everything that distinguishes the two lives off the diagonal.
    \item Off the diagonal, the two behave differently. Since $\sum_{k,k'} P_{k,k'} = 1$ and $\sum_{k,k'} P_k P_{k'} = 1$, we can write $1-P_{k,k} = \sum_{(a,b)\neq(k,k)} P_{a,b}$ and $1-P_k P_k = \sum_{(a,b)\neq(k,k)} P_a P_b$. So, $R_k$ compares the total off-diagonal probability mass with the total mass predicted under independence, giving one lumped number for each $k$. In contrast, the mutual information resolves all $K^2-K$ off-diagonal entries separately. Put plainly, for each cluster $k$, our surprise score asks only a binary question: whether both views are clustered to $k$ more often than independence predicts. In contrast, the mutual information further asks which pair $(k,k')$ the disagreement falls into.
\end{enumerate}

Consequently, the residual $\sum_k R_k - \sum_{k\neq k'} P_{k,k'}\log\left(P_{k,k'}/P_k P_{k'}\right)$ is itself a function of $P$, and hence of $\theta$. The two training targets are therefore not equal, and they are not equal up to an additive constant either. Maximizing one is not the same as maximizing the other, although the two coincide on the diagonal.

\section{Convert the surprise score into differentiable formats and train the model by exploitative approaches}\label{IIC and Gumbel}

Appendix \ref{connect_surprise_score_to_IIC} shows that the surprise score can be made differentiable. Here, we take that step twice --- once with a plain softmax and once with a Gumbel-softmax \citep{jang2016categorical} --- and train the model end-to-end by Adam \citep{AdamOptimization}. Then, we compare the obtained results against those from the converge-to-surprise scheme. 

\subsection{The plain softmax approach}
\label{app_plain_softmax}

Recall the construction of Appendix \ref{connect_surprise_score_to_IIC}. Each argmax of Eq.~\ref{eq:argmax_logit} is replaced by a softmax at temperature $\tau'>0$, so that the two views of image $n$ produce soft assignments on the probability simplex,
\begin{equation}\label{eq:app_plain_assign}
    \boldsymbol{\pi}_n = \mathrm{softmax}\!\left(\mathbf{y}_n^{(i)} / \tau'\right),
    \qquad
    \boldsymbol{\rho}_n = \mathrm{softmax}\!\left(\mathbf{y}_n^{(j)} / \tau'\right).
\end{equation}
Then, every count of Section~\ref{sec:surprise} becomes a soft expectation,
\begin{equation}\label{eq:app_plain_counts}
    \hat q_k = \frac{1}{N}\sum_{n=1}^{N} \pi_{n,k}\,\rho_{n,k},
    \qquad
    p_k = \frac{1}{2N}\sum_{n=1}^{N}\left(\pi_{n,k} + \rho_{n,k}\right),
    \qquad
    q_k = p_k^{\,2}.
\end{equation}
Also, the over-matching indicator becomes the smooth gate of Eq.~\ref{eq:smooth_gate}, giving the differentiable objective of Eq.~\ref{eq:soft_surprise},
\begin{equation}\label{eq:app_plain_score}
    \mathcal{S}_{\mathrm{soft}}(\theta;\tau',\varepsilon)
    = \sum\nolimits_{k} \sigma\!\left(\frac{\hat q_k - q_k}{\varepsilon}\right) D\!\left(\hat q_k \,\Vert\, q_k\right).
\end{equation}
We maximize Eq.~\ref{eq:app_plain_score} directly, taking $-\mathcal{S}_{\mathrm{soft}}$ as the loss. We refer to this as the plain softmax approach.

\subsection{The Gumbel-softmax approach}
\label{app_gumbel_softmax}

A Gumbel variable \citep{marques2015distribution} between 0 and 1, denoted as $\mathrm{Gumbel}(0,1)$, is obtained from a uniform variable by
\begin{equation}\label{eq:app_gumbel_sample}
    g = -\log\!\left(-\log u\right), \qquad u \sim \mathrm{Uniform}(0,1).
\end{equation}
Let $\mathbf{g}_n^{(i)}, \mathbf{g}_n^{(j)}$ be drawn i.i.d.\ from $\mathrm{Gumbel}(0,1)$, independently for each image, each cluster and each of the two views, and re-drawn at every step. Then, the assignments of Eq.~\ref{eq:app_plain_assign} are replaced by
\begin{equation}\label{eq:app_gumbel_assign}
    \boldsymbol{\pi}_n = \mathrm{softmax}\!\left(\mathbf{y}_n^{(i)} / \tau' + \mathbf{g}_n^{(i)}\right),
    \qquad
    \boldsymbol{\rho}_n = \mathrm{softmax}\!\left(\mathbf{y}_n^{(j)} / \tau' + \mathbf{g}_n^{(j)}\right).
\end{equation}
Here, the Gumbel noise is added after the logit vector being divided by $\tau'$. Adding it before, i.e.\ $\mathrm{softmax}\bigl((\mathbf{y}_n + \mathbf{g}_n)/\tau'\bigr)$, is the more popular form. But that is inappropriate in our case: we apply $\ell_2$-normalization to the output logit vector $\mathbf{y}_n$. So, $\mathbf{y}_n$ has entries of order $1/\sqrt{K}$ with measured standard deviation $0.12$, while $\mathrm{Gumbel}(0,1)$ has standard deviation $1.28$. The Gumbel noise will then overwhelm the logit signal by an order of magnitude, making the training process extremely unstable and not reliable. As a result, we add the Gumbel noise after the logit being divided by $\tau'$. 

Everything downstream is unchanged: Eq.~\ref{eq:app_plain_counts} and Eq.~\ref{eq:app_plain_score} are applied to these $\boldsymbol{\pi}_n, \boldsymbol{\rho}_n$ exactly as in the plain softmax approach. That is, dropping $\mathbf{g}$ recovers the plain softmax approach.

By Definition \ref{def_1}, both approaches are exploitative: the parameter updates come from one gradient step, and no random factor is retained by performance-based selection.

\subsection{Results}
\label{app_diff_results}

Both approaches use the same chessboard masking, the same independent augmentations (with the same hyper-parameter settings as described in Table \ref{tab:aug-params} of Appendix \ref{appen_impl}), the same ResNet-9 network backbone with $K=64$, and the same batch size $N$ on each dataset as our main experiments. Only the training objective and the optimizer differ. We set $\tau'=0.02$ and $\varepsilon=0.005$, and use Adam \citep{AdamOptimization} as the training optimizer with a learning rate $10^{-3}$. Once again, $\tau'=0.02$ is the temperature in the softmax Eq.~\ref{eq:app_plain_assign} and Eq.~\ref{eq:app_gumbel_assign}. One should not confuse it with the surprising cluster threshold $\tau$ used in formula \ref{eq:contributing}, whose value is $\tau=0.005$ in our main experiments.

Then, we train the model for $300$ epochs ($900$ on USPS, whose far smaller training set gives correspondingly fewer gradient steps per epoch). End-to-end training is much more efficient than evolution strategies. As a result, we intuitively regard one end-to-end training epoch as equivalent to ten evolution epochs. Thus, we use 300 training epochs here for MNIST and FashionMNIST, corresponding to the 3000 total epochs on these two datasets in the main paper. For the same reason, we use 900 epochs for USPS, corresponding to its 9000 epochs in the main paper. Here, we report the mean $\pm$ standard deviation over five independent runs from fresh initializations.

\begin{table*}[h]
    \centering
    \small
    \setlength{\tabcolsep}{1.0pt}
    \caption{Differentiable (exploitative) versions of the surprise score against the converge-to-surprise scheme (\%, mean $\pm$ std over five runs). All three rows share the identical data pipeline, network architecture and augmentation hyper-parameters. They differ only in the training objective and optimizer. The best mean per column is in \textbf{bold}.}
    \label{tab:differentiable}
    \begin{tabular}{l ccc ccc ccc}
        \toprule
        & \multicolumn{3}{c}{MNIST} & \multicolumn{3}{c}{Fashion-MNIST} & \multicolumn{3}{c}{USPS} \\
        \cmidrule(lr){2-4}\cmidrule(lr){5-7}\cmidrule(lr){8-10}
        Approach & NMI & ARI & ACC & NMI & ARI & ACC & NMI & ARI & ACC \\
        \midrule
        Plain softmax   & $91.3_{\pm3.7}$ & $81.0_{\pm9.0}$ & $82.8_{\pm8.4}$ & $60.6_{\pm3.5}$ & $46.5_{\pm4.4}$ & $58.5_{\pm6.2}$ & $80.4_{\pm4.3}$ & $70.6_{\pm9.0}$ & $72.1_{\pm8.1}$ \\
        Gumbel softmax  & $93.3_{\pm2.0}$ & $87.6_{\pm6.0}$ & $88.2_{\pm7.4}$ & $\mathbf{64.3_{\pm2.3}}$ & $50.2_{\pm3.9}$ & $60.5_{\pm4.2}$ & $84.4_{\pm5.3}$ & $77.0_{\pm9.7}$ & $82.7_{\pm8.3}$ \\
        \midrule
        conv.-to-surprise & $\mathbf{95.6_{\pm0.7}}$ & $\mathbf{96.1_{\pm1.0}}$ & $\mathbf{97.9_{\pm1.3}}$ & $64.0_{\pm1.7}$ & $\mathbf{51.9_{\pm2.2}}$ & $\mathbf{64.1_{\pm1.8}}$ & $\mathbf{89.4_{\pm1.7}}$ & $\mathbf{89.1_{\pm3.0}}$ & $\mathbf{93.4_{\pm3.1}}$ \\
        \bottomrule
    \end{tabular}
\end{table*}

Table~\ref{tab:differentiable} reports the comparison. Both exploitative approaches do learn: the surprise score can be optimized to a good extent by gradient descent. However, neither approach matches the converge-to-surprise scheme applying evolution strategy, which leads on eight of the nine columns --- by $9.7$ ACC points on MNIST, $3.6$ on Fashion-MNIST and $10.7$ on USPS. The only exception is the NMI on Fashion-MNIST, where the Gumbel-softmax is $0.3$ points higher. 

Furthermore, we observe the number of active clusters discovered by both exploitative approaches. We find that granularity is one major shortfall of both exploitative approaches, which under-split the datasets: they discover $8.6$--$9.8$ active clusters on MNIST and Fashion-MNIST, while $7.4$--$9.4$ on USPS. This is obviously lower than $10.1$, $11.2$ and $10.2$ by our converge-to-surprise scheme. 

Comparing to the plain softmax approach, the Gumbel-softmax is better everywhere, and most clearly on USPS ($+10.6$ ACC). Here is our explanation: the sampling noise of Gumbel-softmax perturbs the assignments at every step, keeping the model from settling early into a coarse partition. We regard this as a weak, undirected substitute for mutation-selection. 

Both exploitative approaches are also less stable, with ACC standard deviations of $4$--$8$ points, against $1.3$--$3.1$ from converge-to-surprise scheme. In general, the explorative approach beats exploitative approaches on both performance and stability. 

Finally, we find that in each independent run on each dataset, the differentiable surprise score stops growing in the last $20\%$ epochs. Also, the experimental scores produced by model checkpoints from $80\%$ epochs show no statistical difference comparing to those produced by fully trained models. Hence, 300 epochs on MNIST and Fashion-MNIST as well as 900 epochs on USPS are enough for the training of both exploitative approaches.

Although not a strict proof, experimental results in this appendix section are supportive to Hypothesis \ref{hypothesis_1}. That is, exploitative approaches may optimize a surprise score to some extent, but not fully. A full optimization on the surprise score likely requires explorative approaches, or the mutation-selection mechanism.

\section{Further discussions}
\label{appen_c}

The main paper presents converge-to-surprise as a novel self-supervised learning framework. Here, we step back and discuss the broader picture behind it. We first evaluate the surprise score through the lens of alignment and uniformity, a pair of properties shared by self-supervised objectives in general (Section~\ref{appen_align_uniform}). Then, we argue that the null hypothesis $\mathcal{H}_0$ need not stay fixed: learning can be viewed as a repeated cycle in which the model rejects the current null hypothesis, folds what it discovered into an updated one, and searches again (Section~\ref{appen_iterative}). Finally, we further evaluate the surprise score from an information theory point of view, showing that learning means to build order out of chaos (Section~\ref{appen_orderliness}).

These discussions are conceptual and supplemental. They are not required for the results in the main paper.

\subsection{Alignment and uniformity}
\label{appen_align_uniform}

\citep{wang2020understanding} identified two properties that a contrastive objective such as SimCLR \citep{chen2020simple} optimizes at once: \emph{alignment} pulls the two views of one image together in the representation space, and \emph{uniformity} spreads over the representations of different images as much as possible. Both are necessary, since alignment alone is satisfied by a constant encoder --- that is, by representation collapse. Recently, \citep{mikulasch2026understanding} extended this reading to self-supervised approaches in general.

Our surprise score admits the same reading, with the two properties appearing as the two arguments of the binary Kullback--Leibler divergence in Eq.~\ref{eq:surprise}. The observed view-matching rate $\hat q_k$ is the alignment term, growing only when both views of one image fall into cluster $k$. The marginal clustering rate $p_k$ is the uniformity term, where the distribution $\{p_k\}_{k=1}^K$ needs to spread over all $K$ dimensions in order to minimize the null view-matching rate $q_k={p_k}^2$ for as many clusters as possible. The per-cluster surprise score $D(\hat q_k \,\Vert\, q_k)$ will increase if and only if both the alignment and uniformity properties are satisfied.

As we mentioned in Section \ref{sec:surprise}, our surprise score $\mathcal{S}(\theta)$ naturally prevents the output representation from collapsing. From another perspective, this indicates that neither alignment nor uniformity itself can guarantee effective learning. If the model assigns every image to a single cluster (high alignment alone), then $p_1 = \hat q_1 = q_1 = 1$ and hence $\mathcal{S}(\theta) = 0$. On the other hand, if the clustering of both views are totally independent (high uniformity alone, which also means that the null hypothesis is really true), then there is no significant view-matching at all and hence $\mathcal{S}(\theta) \approx 0$.

\subsection{Learning as iterative rejection of the null hypothesis}
\label{appen_iterative}

In the main paper, without any prior knowledge, we invoked the Principle of Maximum Entropy and assumed the data to be pure random noise. In our case, this means that every pixel is i.i.d. This gives us the null hypothesis $\mathcal{H}_0$. The purpose of our learning framework is then to discover \emph{non-randomness}, or \emph{surprise}, that rejects $\mathcal{H}_0$. However, this is only the starting point of a more general picture.

Once the model has discovered some non-randomness, we may fold that structure back into our description of the data. Concretely, we update the data distribution to account for what is newly found, obtaining an improved hypothesis $\mathcal{H}_1$ that is no longer pure noise: it already encodes the regularities discovered so far. We may now treat $\mathcal{H}_1$ as the new null hypothesis and repeat the process: further evolving and training the network to discover new surprise that rejects $\mathcal{H}_1$. If we succeed, we update the hypothesis again to $\mathcal{H}_2$, take it as the newer null hypothesis, and continue:
\begin{equation}
    \mathcal{H}_0 \;\rightarrow\; \mathcal{H}_1 \;\rightarrow\; \mathcal{H}_2 \;\rightarrow\; \cdots
\end{equation}
Each round explains more of the data, and each new null hypothesis is harder to reject than the last. The process terminates when no further surprise can be found, no matter what output representation we use or how we observe it. At that point, we accept the current hypothesis and stop the learning process.

In general, this refers to the meaning of `learning', which we summarize as three principles:
\begin{enumerate}[leftmargin=*]
    \item \textbf{Maximum-entropy prior.} Without any prior knowledge, we must assume the data distribution to be totally random noise, following the Principle of Maximum Entropy. This is our initial null hypothesis.
    \item \textbf{Learning as discovering surprise.} To learn is to discover \emph{non-randomness}, or \emph{surprise}, from the data distribution in order to reject the null hypothesis.
    \item \textbf{Iterative updating.} We update the null hypothesis to incorporate what we have discovered. Then, we seek new non-randomness (new surprise) that can reject the updated hypothesis. If, after enough updates, no representation and no way of observing can reveal further surprise, we accept the current hypothesis and stop learning.
\end{enumerate}
The converge-to-surprise framework in the main paper realizes a single round of this process. Fully iterating through this process is a natural extension of our framework.

\subsection{Orderliness: building order out of chaos}
\label{appen_orderliness}

Now, we evaluate a surprise score from an information theory point of view. Recall that in Section~\ref{sec:surprise}, we obtain the two cluster sequences from a batch of $N$ images:
$\mathrm{seq}^{(i)} = (\hat y_1^{(i)},\dots,\hat y_N^{(i)})$ and $\mathrm{seq}^{(j)} = (\hat y_1^{(j)},\dots,\hat y_N^{(j)})$.
Pairing the two views of each image gives the cluster pair sequence
\begin{equation}
    \Pi \;=\; \bigl( (\hat y_1^{(i)}, \hat y_1^{(j)}),\, \dots,\, (\hat y_N^{(i)}, \hat y_N^{(j)}) \bigr),
\end{equation}
whose observed distribution over the $K \times K$ cluster pair space is
\begin{equation}
    \pi_{a,b} \;=\; \frac{1}{N}\sum_{n=1}^{N} \mathbf{1}\!\left[\hat y_n^{(i)} = a \ \text{and}\ \hat y_n^{(j)} = b\right].
\end{equation}

Then, we define the \emph{observed pair entropy} as
\begin{equation}
    \hat H_\Pi \;=\; -\!\!\sum_{a,b:\,\pi_{a,b}>0}\!\! \pi_{a,b}\,\log \pi_{a,b}.
\end{equation}

Suppose $n_a^{(i)}$ is the number of times cluster $a$ appears in $\mathrm{seq}^{(i)}$, and $n_b^{(j)}$ is the number of times cluster $b$ appears in $\mathrm{seq}^{(j)}$. Then, it is easy to see that
\begin{equation}
p_a^{(i)} = \sum_b \pi_{a,b}=n_a^{(i)}\ /\ N,\ \ \
p_b^{(j)} = \sum_a \pi_{a,b}=n_b^{(j)}\ /\ N.
\end{equation}
That is, unlike Section \ref{sec:surprise}, we observe the marginal probability distribution of a cluster within each sequence.

Then, consider an inner-view shuffling: we can permute the entries within $\mathrm{seq}^{(i)}$ and within $\mathrm{seq}^{(j)}$ independently, as many times as we like. But $\mathrm{seq}^{(i)}$ and $\mathrm{seq}^{(j)}$ do not exchange entries. Such permutation leaves $\{p_k^{(i)}\}_{k=0}^{K-1}$ and $\{p_k^{(j)}\}_{k=0}^{K-1}$ unchanged, but scrambles the pairing and hence the cluster pair sequence $\Pi$. 

Given the original $\Pi$, suppose all pair sequences that we can possibly obtain after inner-view shuffling form the set $\mathcal{R}_{\Pi}$. Then, we define the \emph{shuffled-maximal entropy} as the largest attainable value in the form:
\begin{equation}
    H_\Pi^{\star} = \max_{\tilde{\Pi}\in \mathcal{R}_{\Pi}}\hat H_{\tilde{\Pi}}.
\end{equation}
That is, we calculate the observed pair entropy $\hat H_{\tilde{\Pi}}$ for each pair sequence $\tilde{\Pi}\in \mathcal{R}_{\Pi}$, and then take the maximum as $H_\Pi^{\star}$.

It is easy to see that $\hat H_\Pi \le H_\Pi^{\star}$ always holds true. Also, by a routine analysis, we can get that
\begin{equation}
    H_\Pi^{\star} = H\!\left(p^{(i)}\right) + H\!\left(p^{(j)}\right),
    \ \
    H(p^{(\cdot)}) = -\sum_{k} p_k^{(\cdot)} \log p_k^{(\cdot)} ;
\end{equation}
and
\begin{equation}
    H_\Pi^{\star} - \hat H_\Pi \;=\; I\!\left(\hat y^{(i)}; \hat y^{(j)}\right) \;\ge\; 0.
\end{equation}
Here, $I\!\left(\hat y^{(i)}; \hat y^{(j)}\right)$ is the mutual information between the original, un-shuffled $\mathrm{seq}^{(i)}$ and $\mathrm{seq}^{(j)}$, as described in Section \ref{sec:surprise}.

Finally, we define the \emph{orderliness} (or \emph{neg-entropy}):
\begin{equation}
    \Omega \;=\; \frac{H_\Pi^{\star} + \lambda}{\hat H_\Pi + \lambda},
    \label{eq:orderliness}
\end{equation}
where $\lambda > 0$ is a small constant that keeps the ratio well-defined in degenerated cases. Since $\hat H_\Pi \le H_\Pi^{\star}$, we always have $\Omega \ge 1$, with larger values indicating more order.

Then, we describe two special cases:
\begin{enumerate}[leftmargin=*]
\item \textbf{Fresh initialization:} Right after initializing the model, $\mathrm{seq}^{(i)}$ and $\mathrm{seq}^{(j)}$ are essentially independent. So, the observed pairing is already close to its most disordered form. Thus, we have $\hat H_\Pi \approx H_\Pi^{\star}$ and $\Omega \approx 1$. This matches $\mathcal{S}(\theta)\approx 0$: nothing surprising has been found.
\item \textbf{Collapse:} If the network collapses so that every view is mapped to a single cluster $k_0$, then only the pair $(k_0,k_0)$ is ever observed. In this case, no shuffling can change the cluster pair sequence $\Pi$. Hence, $H_\Pi^{\star} = \hat H_\Pi = 0$ and $\Omega = \lambda/\lambda = 1$ always hold true. In other words, collapse is not learning.\\
\end{enumerate}
Thus, genuine learning drives $\Omega \gg 1$, which by Eq.~\ref{eq:orderliness} requires the observed pair entropy $\hat H_\Pi$ to be much smaller than the shuffled-maximal entropy $H_\Pi^{\star}$. The latter demands diverse marginals (many clusters being observed), i.e.\ high potential disorder or high uniformity; the former demands the actual pairing to be highly dependent (both views of an image tend to agree), i.e.\ strong order or strong alignment. In other words, to learn is to realize uniformity and alignment at the same time, or to keep the marginal diversity while imposing order on how these diversities are organized --- to \emph{build order out of chaos}.

Combining this with Section~\ref{appen_iterative}, we arrive at an intuitive picture of the whole process: learning repeatedly discovers order out of chaos and folds what being discovered into an ever-stronger hypothesis, and continues until there is no more order left to discover.

\section{The algorithm of converge-to-surprise scheme}
\label{CtS Algorithm}
Algorithm \ref{alg:cts} shows the complete converge-to-surprise scheme:
\begin{algorithm}[h]
\caption{Converge to Surprise}
\label{alg:cts}
\begin{algorithmic}[1]
\Require images $\mathcal{X}$; population $m$; noise $\sigma$; learning rate $\eta$; weight decay $\lambda$; warm-up $T_0$; training periods $T_{1,a}$ and $T_{1,b}$; threshold $\tau$.
\State initialize $\theta$
\For{epoch $e = 1, 2, \dots$}
    \State sample $N$ images; build complementary masked views and augment each independently $\rightarrow \tilde{\mathbf{X}}^{(i)}, \tilde{\mathbf{X}}^{(j)}$
    \Statex \textit{\quad // ES outer step (explore)}
    \For{$p = 1$ to $m/2$}
        \State draw $\epsilon_p \sim \mathcal{N}(0, \sigma^2 I)$
        \State score $\theta \pm \epsilon_p$ using $\mathcal{S}(\cdot)$ (Eq.~\ref{eq:surprise})
    \EndFor
    \State shape the $m$ scores into centered ranks $r_p$
    \State $\theta \gets (1-\eta\lambda)\,\theta + \frac{\eta}{m\sigma}\sum_p r_p\,\epsilon_p$ \Comment{Eq.~\ref{eq:es-update}}
    \Statex \textit{\quad // gradient-descent training (consolidate)}
    \If{$e > T_0$ and $e \mathbin{\%} T_{1,b} = 0$}
        \State on un-augmented views, take argmax labels and keep contributing positions $\mathcal{C}$ (Eq.~\ref{eq:contributing})
        \State median-balance $\mathcal{C}$ across clusters into $\mathcal{B}$
        \For{epoch $e = 1, 2, \dots, T_{1,a}$}
            \State re-augment both views; update $\theta$ by minimizing $\mathcal{L}_\text{ft}(\theta)$ (cross-entropy to the surrogate labels)
        \EndFor
    \EndIf
\EndFor
\State \Return $\theta$
\end{algorithmic}
\end{algorithm}

In summary, when the optimization begins, we do not have a clear training loss in each step to exploit. So, we just let the model explore and evolve toward the ultimate goal (the surprise score). After enough epochs, we may see some promising output representations, which in our case are the clusters with surprising view-matching results. Then, we train the network using these clusters as surrogate labels to further strengthen, or reinforce, the discovered promising representations.

\section{Implementation details}\label{appen_impl}

\textbf{Network:}
Our model $f_\theta$ is a ResNet-9 \citep{he2016deep_residual_learning}: two convolutional stems \citep{aloysius2017review}, two residual blocks \citep{he2016deep_residual_learning}, an adaptive average-pooling layer, and a single linear head that returns a logit vector with dimension $K=64$ . The cluster assignment of a view is the argmax over these logits. The output is $\ell_2$-normalized before the argmax. We set $K=64$, far above the ten ground-truth classes of every dataset. Since the model is never told the number of true classes, $K$ only acts as an upper bound. The number of active clusters is discovered during training. The same architecture is used for all datasets, with a single input channel for an image view.

\textbf{Datasets:}
We use three handwritten-digit and fashion-item benchmarks, all with ten ground-truth classes. \textbf{MNIST} \citep{MNIST_paper} contains 70k images of handwritten digits at $28\times28$ resolution (10k for testing). \textbf{Fashion-MNIST} \citep{xiao2017fashion} matches MNIST in size and resolution but depicts ten clothing categories, and is substantially harder because several classes (e.g.\ pullover/coat/shirt) differ only in fine texture. \textbf{USPS} \citep{hull2002database} contains $7{,}291$ training and $2{,}007$ test handwritten digits at a native $16\times16$ resolution, and is notably class-imbalanced. %After training, our model is evaluated on the test split against the ground-truth labels.

\textbf{Data augmentation:}
After the chessboard masking of Section~\ref{sec:masking}, the two complementary views of each image are augmented independently, preserving the zero-mutual-information property between two views under $\mathcal{H}_0$. For MNIST and Fashion-MNIST, augmentations include random rotation, random cropping, and brightness/contrast jitter. Fashion-MNIST additionally applies a random horizontal flip with probability $0.5$. Digits are not flipped, as a mirrored digit is not the same class. 

For USPS, we add a single-axis zoom-out after cropping, which pads one randomly chosen axis with a black border before resizing back, producing an anisotropic (tall/short, wide/narrow) deformation. This is motivated by the imbalance of the USPS dataset: the digit-0 class is dominant and exhibits far more aspect-ratio variation than its MNIST counterpart. The anisotropic augmentation de-correlates this global shape so that same digits are less likely to be further split based on width or height. To leave room for the anisotropic deformation, the remaining augmentations on USPS images are kept milder than on MNIST/Fashion-MNIST images. Besides, USPS characters occupy a relatively larger fraction of the frame. So, we mask at the native $16\times16$ resolution, upsample each view independently to $32\times32$, and crop only down to a $30\times30$ window (i.e., a milder cropping range: $30\sim 32$). Table~\ref{tab:aug-params} lists the exact per-dataset augmentation parameters referenced in Section~\ref{sec:exp-setup}.
\begin{table}[h]
    \centering
    \small
    \setlength{\tabcolsep}{5pt}
    \caption{Per-dataset augmentation parameters, applied independently to each view after chessboard masking. Rotation angle is drawn uniformly from $[-r, r]$, and being applied with a rotation probability. Brightness/contrast are multiplicative jitters drawn from $[1-b,\,1+b]$. The square cropping size (cropped edge length) is sampled uniformly from the listed range and resized back to $S\times S$. Also, edge length $S$ is among the listed range of cropping. This means that although the cropping probability is set to 1, there is actually a chance of sampling $S$ itself and implementing no cropping. Since USPS digits are relatively larger, the listed range of cropping is narrower. Also, for USPS, we first obtain the masked views from the original $16\times16$ image. Then, we upsample per view to $32\times32$. Additionally, we pad one random axis with $2$--$6$ black pixels (zoom-out) independently on each view, which is only applied to USPS. A `--' denotes a disabled augmentation.}
    \label{tab:aug-params}
    \begin{tabular}{l ccc}
        \toprule
        Parameter & MNIST & FMNIST & USPS \\
        \midrule
        Side $S$                 & $28$         & $28$         & $32$ \\
        Rotation $r$ (deg)       & $20$         & $20$         & $10$ \\
        Rotation prob           & $1.0$        & $1.0$        & $0.5$ \\
        Bright/contrast $b$     & $0.3$        & $0.3$        & $0.15$ \\
        Crop edge                & $\{24..28\}$ & $\{24..28\}$ & $\{30..32\}$ \\
        Crop prob               & $1.0$        & $1.0$        & $1.0$ \\
        Flip prob               & --           & $0.5$        & -- \\
        Zoom-out prob           & --           & --           & $1.0$ \\
        Zoom-out range           & --           & --           & $\{2..6\}$ \\
        \bottomrule
    \end{tabular}
\end{table}

\textbf{Optimization and protocol:}
We optimize $\mathcal{S}(\theta)$ with the ES outer loop under a population size $m=32$, interleaved with periodic gradient-descent training using the discovered surprising clusters. When identifying contributing positions according to formula \ref{eq:contributing}, we choose $\tau=0.005$. For each dataset, we run five independent experiments from scratch and report the mean $\pm$ standard deviation. At test time, only view-i (without augmentation) from chessboard masking on each test image is passed through the network, and assigned to its argmax cluster. The clustering results are then used in evaluations.

\textbf{Training schedule:}
On MNIST and Fashion-MNIST, we use a two-stage training schedule: $2000$ epochs of pure ES, followed by epochs $2000$--$3000$ in which $4$ gradient-descent training epochs are run every $25$ ES epochs. On USPS we use a three-stage schedule: $4000$ epochs of pure ES; then a weak stage over epochs $4000$--$8000$, with $2$ gradient-descent training epochs every $500$ ES epochs; and finally a strong stage over epochs $8000$--$9000$, with $4$ gradient-descent training epochs every $25$ ES epochs (identical to the other datasets). USPS is run for many more epochs because it has far fewer training images ($7{,}291$ vs.\ $70{,}000$), so the model needs more epochs to evolve.

\textbf{Batch size:}
The surprise score is calculated over a batch of $N$ images at each ES step. We use $N=3000$ for MNIST/Fashion-MNIST, and $N=3650$ for USPS. The latter is chosen so that the $7{,}291$ USPS training images are split into two nearly equal-sized steps per epoch ($3650$ and $3641$). This matters because each ES updating direction is estimated from the statistics of its batch alone. Then, a last step with only a few images gives unreliable statistics, steering the evolution into a poorly-estimated direction. So, we choose $N$ carefully to avoid this.

\textbf{Metrics and baselines:}
Following the non-parametric clustering protocol, we report three standard metrics: clustering accuracy (ACC) under the Kuhn-Munkres (Hungarian) algorithm \citep{kuhn1955hungarian}, Normalized Mutual Information (NMI) \citep{kvaalseth2017normalized}, and the Adjusted Rand Index (ARI) \citep{santos2009use}. For all three, higher is better, with more details introduced in \citep{ronen2022deepdpm}. We additionally report the inferred number of active clusters $\hat K$ (ground truth $K^\star=10$). As for baselines, we use the scores reported by DeepDPM \citep{ronen2022deepdpm} --- including the performances of DeepDPM itself, performances of the classical non-parametric clusterers DBSCAN \citep{ester1996density}, performances of a memorized variational DPM model called moVB \citep{hughes2013memoized}, and performances of a DPM sampler. According to DeepDPM \citep{ronen2022deepdpm}, all these models are based on pre-trained autoencoder features. Also, performances of the three UNSEEN variants \citep{leiber2024dying} (UNSEEN+DCN \citep{yang2017towards}, UNSEEN+DEC \citep{xie2016unsupervised}, and UNSEEN+DKM \citep{fard2020deep}), each of which is also based on a pre-trained autoencoder backbone, are used as baselines. In contrast to all these, our method is trained from scratch, with no pre-trained or separately-learned feature extractor.

\textbf{Computation and hardware:}
All our experiments are conducted using one NVIDIA H200 GPU. We apply parallel computing within a single GPU: Given the ES population size $m=32$, we duplicate the network parameters for 16 times, perturb each of them individually, and implement them on 16 duplicated batches of image views. So, one ES step requires $32/16=2$ iterations, and in each iteration there are $16/2=8$ mirrored pairs. It takes around 5 hours to fully optimize a ResNet-9 model on MNIST or FashionMNIST, and around 2 hours on USPS.

\section{Number of discovered clusters during training}\label{appen_num_clusters}

\begin{figure}[h]
    \centering
    \includegraphics[width=0.7\linewidth]{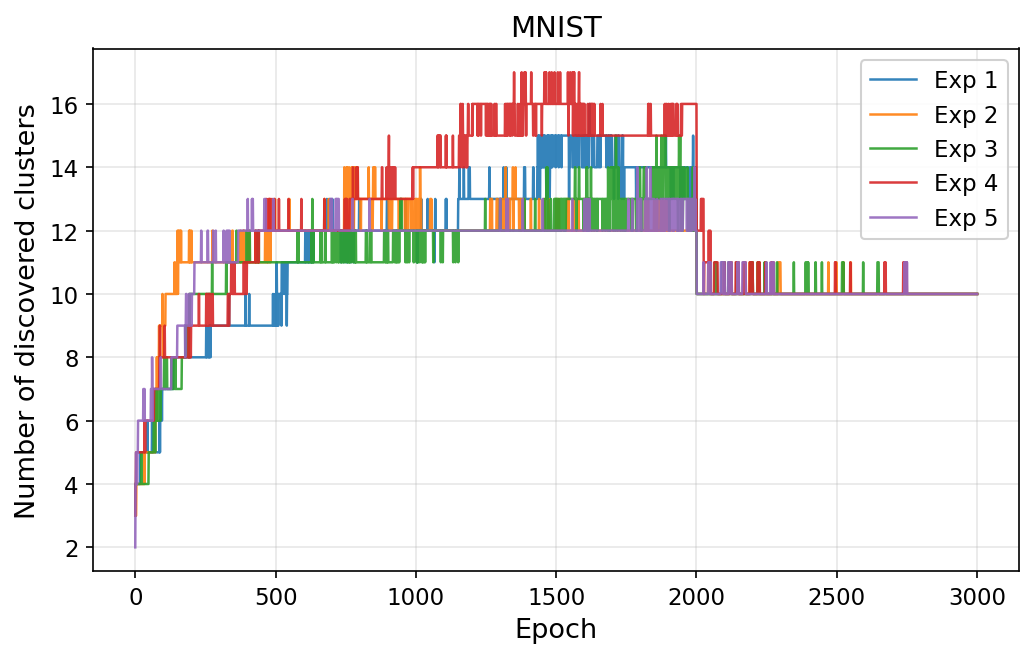}
    \caption{Number of discovered (surprising) clusters vs.\ training epoch on MNIST, for five independent runs. The model over-produces surprising clusters during pure ES stage. Then, the number of surprise clusters are consolidated back to $10$ once gradient-descent training begins (two-stage schedule, starts from epoch $2000$).}
    \label{fig:nc-mnist}
\end{figure}

\begin{figure}[h]
    \centering
    \includegraphics[width=0.7\linewidth]{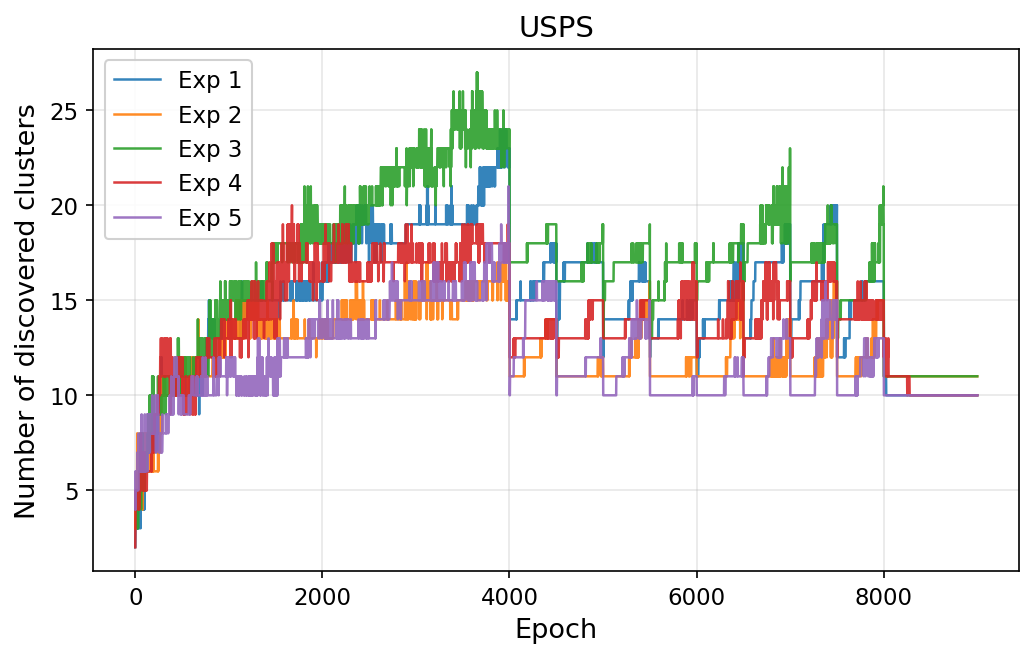}
    \caption{Number of discovered (surprising) clusters vs.\ training epoch on USPS, for five independent runs (three-stage schedule). Again, pure ES stage ($0$--$4000$) over-produces; weak training stage ($4000$--$8000$) yields the consolidation/re-exploration sawtooth; the final strong gradient-descent training stage settles the number of surprising clusters in each run to $10$ or $11$.}
    \label{fig:nc-usps}
\end{figure}

\begin{figure}[h]
    \centering
    \includegraphics[width=0.7\linewidth]{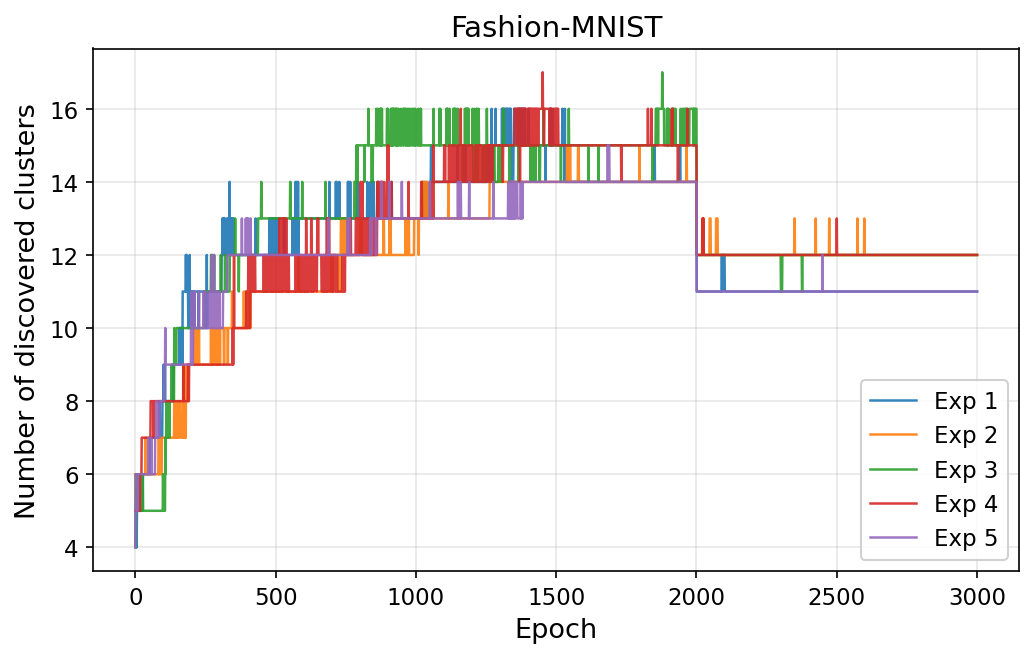}
    \caption{Number of discovered (surprising) clusters vs.\ training epoch on Fashion-MNIST, for five independent runs. The curves behave similarly as in the experiments on MNIST, whereas the final number of surprising clusters is slightly above 10. This is in line with the dataset's harder, texture-dominated structure.}
    \label{fig:nc-fashion}
\end{figure}

Because the number of classes is never given to the model, it is instructive to watch how many clusters the model discovers as training proceeds. Recall that we introduced $D\!\left(\hat q_k \,\Vert\, q_k\right)$, the surprise score at cluster $k$, via formula \ref{surprise_score_k} in Section~\ref{sec:surprise}. Then, we regard cluster $k$ as \emph{surprising} if $D\!\left(\hat q_k \,\Vert\, q_k\right) \geq \tau=0.005$. Figures~\ref{fig:nc-mnist}, \ref{fig:nc-usps} and \ref{fig:nc-fashion} plot, for five independent runs per dataset, the number of surprising clusters after each ES epoch. 

We can see that during the pure-ES stage, the model keeps discovering new candidate clusters, exceeding the number of ground truth classes. We call this over-production. Once gradient-descent training begins, each training round consolidates redundant clusters, pulling the count back down. Then, the evolution strategy slightly rebuilds surprising clusters on the cleaner partition. But finally, the evolution strategy and gradient-descent training reach the balance. No new surprising cluster is discovered in the last 200 epochs. In general, results shown here coincide with our analysis in the ablation study \ref{sec:exp-ablation}.

On MNIST (Figure~\ref{fig:nc-mnist}, two-stage schedule), the over-production peaks in the first $2000$ epochs, and then collapses cleanly to $10$ for all five runs. On USPS (Figure~\ref{fig:nc-usps}, three-stage schedule) the effect is more pronounced: the pure-ES stage ($0$--$4000$ epochs) over-produces up to around $25$ clusters. In the weak stage ($4000$--$8000$ epochs, with 2 training epochs every 500 ES epochs), the ES optimization rebuilds back the number of surprising clusters after every gradient-descent training epoch, creating the sawtooth curves. Finally, the strong stage ($8000$--$9000$ epochs, with 4 training epochs every 25 ES epochs) settles the number of surprising clusters to $10$ or $11$. Fashion-MNIST (Figure~\ref{fig:nc-fashion}) behaves similarly but stabilizes at a slightly larger count, reflecting its harder, texture-dominated data structure. 

Once again, by converge-to-surprise, the model naturally discovers meaningful clusters from raw images without any prior knowledge.

\section{Clustering visualization}\label{appen_visualization}

To qualitatively inspect what the network discovered, we visualize the test-set clustering of one trained model per dataset. For each active cluster $k$ and each ground-truth class $p$, we draw a $3\times3$ grid of randomly selected test images that belong to cluster $k$ and carry label $p$. We only draw such a grid when at least $50$ such images exist (rare classes in each cluster are ignored). Each panel is captioned ``Cluster $k$, digit $p$'', or ``Cluster $k$, item $p$''.

\begin{figure*}[h]
    \centering
    \includegraphics[width=\linewidth]{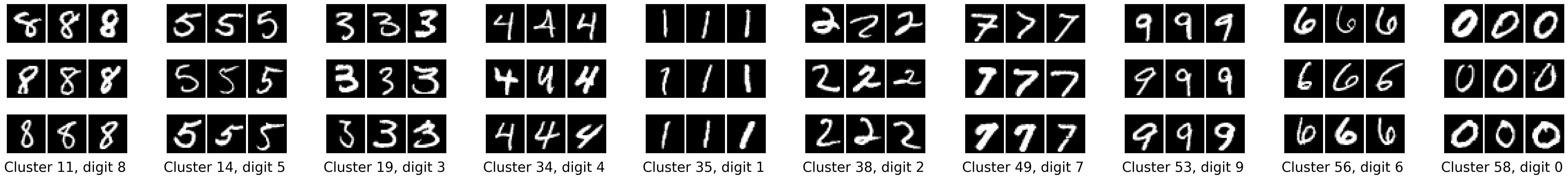}
    \caption{MNIST test-set clustering. Each panel shows nine random images of a single ground-truth class assigned to a given cluster. The ten active clusters map one-to-one to the ten digits, each at $97$--$99.7\%$ purity.}
    \label{fig:vis-mnist}
\end{figure*}

\textbf{MNIST:} Figure~\ref{fig:vis-mnist} shows the MNIST clustering. The optimized network produces a clean one-to-one map: each active cluster corresponds to a single ground-truth class, and each class is captured by a single cluster. This is confirmed by the cluster-purity report, where every active cluster is dominated by one digit class at $97\%$ -- $99.7\%$ purity:
\begin{verbatim}
Cluster 35 | n=1134 | 99.0% (dom=cls1)
Cluster 38 | n=1043 | 98.6% (dom=cls2)
Cluster 19 | n=1036 | 97.0% (dom=cls3)
Cluster 49 | n=1031 | 98.5% (dom=cls7)
Cluster 53 | n= 997 | 97.1% (dom=cls9)
Cluster 58 | n= 980 | 99.4% (dom=cls0)
Cluster 56 | n= 968 | 98.3% (dom=cls6)
Cluster 34 | n= 968 | 98.9% (dom=cls4)
Cluster 11 | n= 958 | 99.7% (dom=cls8)
Cluster 14 | n= 885 | 99.0% (dom=cls5)
\end{verbatim}
Each of the ten digits is represented by exactly one high-purity cluster, so Figure~\ref{fig:vis-mnist} contains exactly ten panels.

\begin{figure*}[h]
    \centering
    \includegraphics[width=\linewidth]{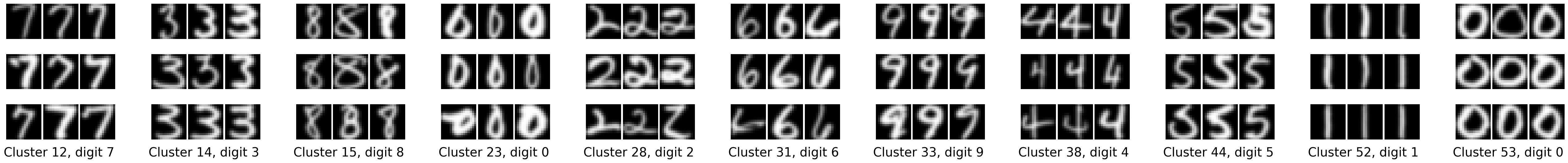}
    \caption{USPS test-set clustering for a run that inferred $11$ clusters. Clusters $23$ and $53$ both capture the digit $0$: cluster $53$'s zeros are round and plain, while cluster $23$'s are narrow, tall, and often carry an extra stroke, which is a residual geometric variation the anisotropic augmentation did not absorb.}
    \label{fig:vis-usps}
\end{figure*}

\textbf{USPS:} Next, Figure~\ref{fig:vis-usps} shows a deliberately less-than-perfect USPS run, in which the model settled on $11$ clusters instead of $10$. Its per-cluster purity is still high, but two clusters share the same dominant class (digit 0), which is what pushes the count to $11$:
\begin{verbatim}
Cluster 53 | n=313 | 97.1% (dom=cls0)
Cluster 52 | n=256 | 99.6% (dom=cls1)
Cluster 28 | n=199 | 95.0% (dom=cls2)
Cluster 38 | n=199 | 94.5% (dom=cls4)
Cluster 33 | n=188 | 93.1% (dom=cls9)
Cluster 15 | n=168 | 95.2% (dom=cls8)
Cluster 14 | n=166 | 97.6% (dom=cls3)
Cluster 31 | n=162 | 98.1% (dom=cls6)
Cluster 44 | n=157 | 95.5% (dom=cls5)
Cluster 12 | n=144 | 96.5% (dom=cls7)
Cluster 23 | n= 55 | 90.9% (dom=cls0)
\end{verbatim}
Both cluster $23$ and cluster $53$ focus on the digit $0$: the zeros in cluster $53$ are relatively round and plain, whereas those in cluster $23$ are narrow, tall, and often carry an extra stroke beyond the bare loop. This shows that the zoom-out (anisotropic) augmentation largely works as intended, but does not perfectly remove every unwanted topological and geometric variation of the dominant digit-0 class. This leads to further split of the digit-0 class.

\begin{figure*}[h]
    \centering
    \includegraphics[width=\linewidth]{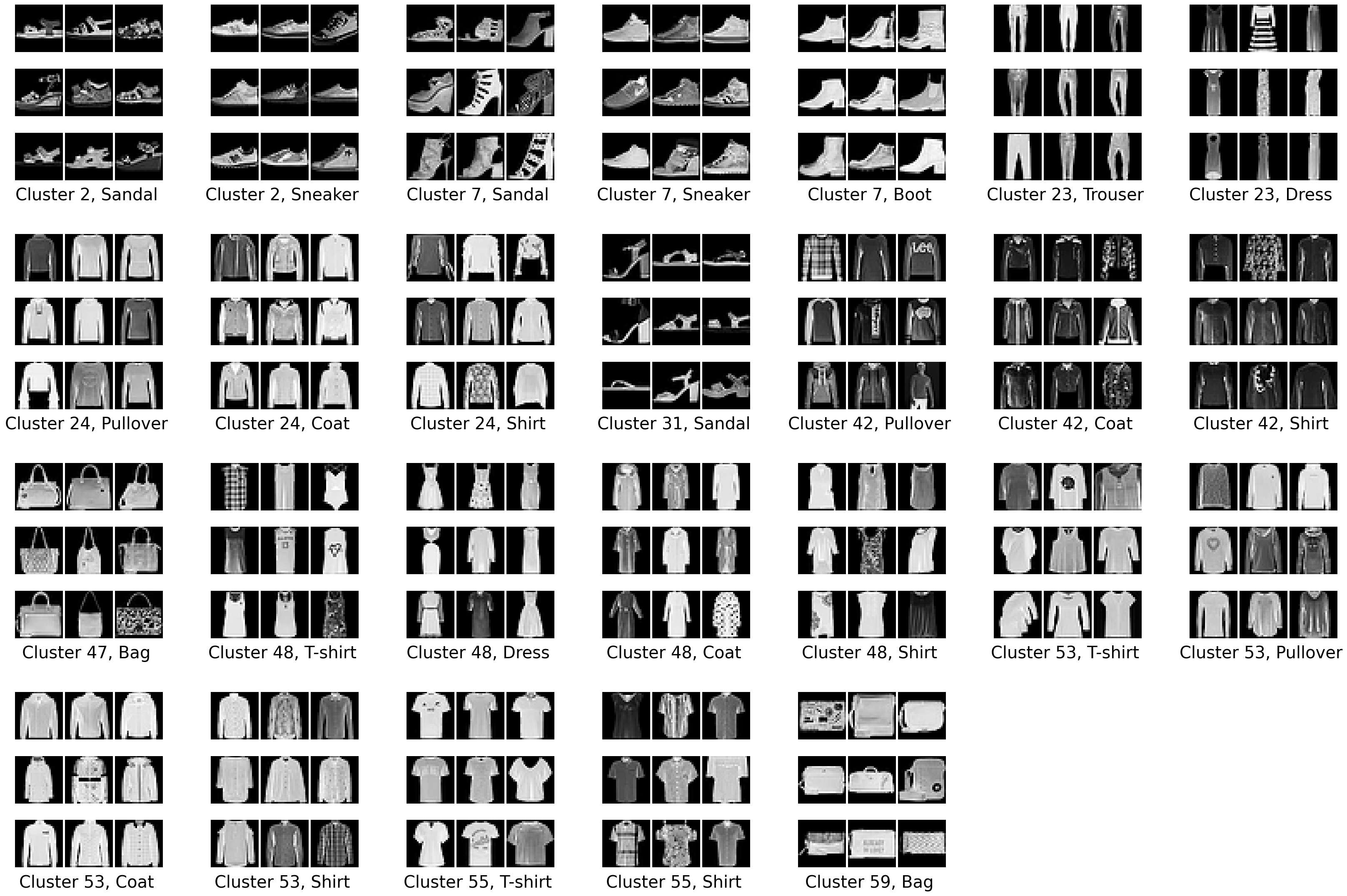}
    \caption{Fashion-MNIST test-set clustering. Beyond coarse categories, the network separates finer attributes: clusters $47$ vs.\ $59$ split bags by the presence of a handle/strap, and clusters $24$ vs.\ $53$ split upper-body garments by texture (patterned vs.\ plain) rather than contour.}
    \label{fig:vis-fashion}
\end{figure*}

\textbf{Fashion-MNIST:} Finally, Fashion-MNIST is far more challenging for non-parametric self-supervised image clustering, since several classes differ only in fine texture rather than global contour \citep{xiao2017fashion}. The per-cluster purity is correspondingly lower and some clusters mix classes:
\begin{verbatim}
Cluster 53 | n=1303 | 31.1% (dom=cls6)
Cluster 23 | n=1165 | 82.1% (dom=cls1)
Cluster 48 | n=1164 | 63.4% (dom=cls3)
Cluster  2 | n=1136 | 81.4% (dom=cls7)
Cluster  7 | n=1112 | 86.2% (dom=cls9)
Cluster 55 | n=1014 | 76.4% (dom=cls0)
Cluster 24 | n= 979 | 46.7% (dom=cls2)
Cluster 31 | n= 770 | 99.2% (dom=cls5)
Cluster 47 | n= 524 | 96.8% (dom=cls8)
Cluster 59 | n= 476 | 90.1% (dom=cls8)
Cluster 42 | n= 357 | 40.9% (dom=cls2)
\end{verbatim}
Even so, the visualization in Figure~\ref{fig:vis-fashion} shows that the network discovers meaningful structure beyond the labels. Clusters $47$ and $59$ both isolate bags, but along a finer distinction: cluster $47$ collects bags with a visible handle/strap, whereas cluster $59$ collects bags with no (or only a very small, barely visible) handle. Clusters $24$ and $53$ both contain upper-body garments, but separate them by texture rather than contour: the garments in cluster $24$ are more heavily textured/patterned, while those in cluster $53$ are comparatively plain. In other words, despite using only a ResNet-9, converge-to-surprise optimization enables the network to distinguish both contour and texture.

Once again, although the per-cluster purity on Fashion-MNIST is lower than those on the other two datasets, we still achieve state-of-the-art performances on Fashion-MNIST.

\section{Failure mode in CIFAR10}\label{fail_CIFAR10}

Our scheme does not transfer successfully to CIFAR10 \citep{krizhevsky2009learning}. We report this
plainly, together with what we tried and what we believe the reasons are. The
contrast between these failure modes and the successful results on digit and garment benchmarks is itself informative about when a
surprise score aligns with semantics.

\subsection{What was tried}
\label{fail_cifar_table}

Every configuration in Table~\ref{tab:cifar-failure} uses the full converge-to-surprise scheme --- an evolution strategy (ES) outer loop with periodic gradient-descent inner loop as described in Section~\ref{sec:optimization} --- with $K=64$ and the same ResNet-9 backbone model (ResNet-34 is tried in one setting). But we make the batch size to be $N=3125$ on CIFAR10, in which case the 50,000 training images can be equally divided into 16 batches each epoch. The major difference occurs in data augmentation: we use different methods to augment the colored views.

\begin{table*}[h]
    \centering
    \small
    \setlength{\tabcolsep}{4pt}
    \caption{Results (in percentage) on CIFAR10 by the converge-to-surprise scheme. Each row applies its unique data augmentation configuration or network backbone. $\hat K$ is the number of active clusters out of $K=64$. The ground truth is $K^\star = 10$. Chance accuracy is $10\%$. We can see that every configuration lands in the same band --- NMI $7$--$12$, ACC $20$--$24$ --- regardless of how the color and texture cues are augmented or modified. Repeated runs of one configuration vary by about one NMI point. So, the differences between rows are of the same order as the noise.}
    \label{tab:cifar-failure}
    \begin{tabular}{l c ccc c}
        \toprule
        View construction & input channels & NMI & ARI & ACC & $\hat K$ \\
        \midrule
        Per-view color tint, ResNet-34         & $3$ & $9.1$  & $5.7$ & $23.0$ & $6$ \\
        Per-view color tint                    & $3$ & $8.8$  & $5.5$ & $22.5$ & $6$ \\
        Greyscale $+$ smoothing, with rotation  & $1$ & $6.9$  & $4.1$ & $20.0$ & $6$ \\
        Greyscale $+$ smoothing, no rotation    & $1$ & $6.7$  & $3.4$ & $19.6$ & $9$ \\
        Sobel, squashed encoding                & $2$ & $11.4$ & $5.6$ & $20.5$ & $12$ \\
        Sobel, IIC-aligned jitter               & $2$ & $\mathbf{12.0}$ & $\mathbf{6.5}$ & $\mathbf{23.7}$ & $10$ \\
        Sobel, IIC-matched crop                 & $2$ & $10.3$ & $5.8$ & $21.4$ & $6$ \\
        \bottomrule
    \end{tabular}
\end{table*}

In row 1 and 2 of Table~\ref{tab:cifar-failure}, color tint gives each view its own additive color offset, so that absolute color cannot agree across views. Row 2 applies ResNet-34 instead of the standard ResNet-9 across our experiments. In row 3 and 4, greyscale removes color outright and fills the chessboard masked pixel positions with the mean of their visible neighbors. Row 4 further cancels rotation in each view. In row 5, 6 and 7, Sobel augmentation \citep{shorten2019survey} replaces each view by its two spatial-gradient channels. Following IIC \citep{ji2019invariant}, row 5 removes absolute brightness as well. Finally, row 6 and 7 additionally match IIC's photometric jitter and its crop aggressiveness.

Table~\ref{tab:cifar-failure} shows that none of this changes the outcome. Every configuration lands between $19.6\%$ and $23.7\%$ accuracy, against $10\%$ of pure guessing and against the $98.3\%$, $64.5\%$ and $94.9\%$ our scheme reaches in Table~\ref{tab:main-results}. Different configurations move NMI by a few points, but never changes the character of the result. For reference, IIC reports $61.7\%$ accuracy (ACC) on CIFAR10 with $K=10$ given to the model. 

In addition, the failure seems not to stem from the converge-to-surprise scheme itself. On the greyscale configuration, the
surprise score reached $1.054$ during pure ES, a value indistinguishable from the digit
runs ($1.21$ on MNIST, $1.17$ on Fashion-MNIST, $1.25$ on USPS), while the mutual
information with the ground-truth labels stayed near $0.07$. That is, the model finds
view-consistent features, which are surprising enough but unfortunately not semantic related.

\subsection{Why it fails}
\label{fail_cifar_why}

\textbf{The evolution strategy applied in this paper is preliminary, which is under-powered for natural images.} Across all our experiments, each ES step estimates a parameter update direction from a population of $m=32$ (with mirrored perturbation) in a parameter space of $107{,}320$ dimensions. A zeroth-order estimate of this kind pays a cost in the ambient dimension that back-propagation does not, and what it can resolve is correspondingly coarse. On handwritten digits that suffices, because a digit essentially is its low-frequency shape, so the direction that improves the surprise score is also the direction that improves the clustering. On natural images, the features that separate a cat from a dog are far more complex than a $32$-sample directional estimate can resolve. Thus, Table \ref{tab:cifar-failure} shows that the parameter space search settles on whatever coarse statistic the two views share. 

The obvious remedy is to enlarge the population substantially. However, to accelerate the optimization process, the entire population is evaluated in parallel in a single fused pass and its memory cost grows with $m$. With a single GPU (although the advanced NVIDIA H200), the memory consumption is unfeasible to us. On the other hand, if we remove the parallel computing and iterate on each mirrored perturbation pair, the total training time explodes. We regard this as a limitation of our approach. In the future, smarter evolution strategies will be explored.

\textbf{The surprise score in this paper is preliminary, which provides insufficient distinguishing among views of different images.} This is the deeper reason. Contrastive learning \citep{chen2020simple} pulls the two views of one image together and explicitly pushes views of different images apart. Self-distillation \citep{caron2021emerging} achieves a comparable effect by centering and sharpening the teacher's output over a very large set of prototypes (output dimensions), so that views can hardly be predicted to the same output distribution. 

Our surprise score does neither. It asks only this: the two views of the same image should agree more often than what the independence baseline $q_k = p_k^{\,2}$ predicts. Once any output cluster achieves any kind of surprising view-matching results, the objective is satisfied. Nothing else forces two different images to be set apart. With only $K=64$ output dimensions and no such pressure, a partition that merely reflects a low-level cue shared by both views --- global brightness layout, or edge-density layout after Sobel --- scores as well as a semantic one. On digits or garment items this is harmless, since the easiest view-consistent feature happens to be the class. On natural images, there is a whole ladder of view-consistency. Non-semantic features occupy the low level and are much easier to capture. Hence, the evolution never manages to climb the ladder to reach at the semantic level.

Still, some perspective is due. Again, to the edge of our knowledge, our converge-to-surprise scheme is the first explorative SSL approach trained from scratch for self-supervised image clustering. Proposing a novel self-supervised learning framework is never easy, and making it scale to every dataset at the first attempt is even harder. We claim that the failure on CIFAR10 comes from two technical deficiencies --- a surprise score that is too sparse, and an evolution strategy that is too naive --- rather than from the converge-to-surprise principle itself.

Both issues have drawn our attention, and both point to concrete next steps. On the objective, we will study surprise scores over denser output representations: several heads at different granularities, or a factorized code whose number of states far exceeds $K=64$. On the optimizer, we will replace the fixed perturbation scale by step-size and covariance adaptation \citep{wierstra2014natural}, and add an acceptance test, so that one generation cannot move the parameters when every sampled candidate scores worse than the current ones. 

Anyway, the new state-of-the-art performances on the digit and garment benchmarks show the effectiveness of the converge-to-surprise principle. Extending that effectiveness to natural images is our next step, rather than a refutation of the framework.

\end{document}